\providecommand{\TechnicalReport}{1}
\documentclass[final,3p,times]{elsarticle}
\newcommand{\ReportNote}{\ifnum\TechnicalReport=1\relax\space Technical report.\fi}
\usepackage[T1]{fontenc}
\usepackage{amsmath,amssymb,booktabs,listings,tabularx,xcolor}
\usepackage[scaled=0.95]{inconsolata}
\usepackage{xurl}
\usepackage[colorlinks=true,breaklinks=true]{hyperref}
\usepackage{tikz}
\usepackage{placeins}
\usetikzlibrary{arrows.meta}

\definecolor{InterfaceBlue}{HTML}{005FBB}
\definecolor{InterfaceTeal}{HTML}{007858}
\definecolor{CodeInk}{HTML}{202733}
\definecolor{CodeRule}{HTML}{A8B2BC}
\definecolor{CodeComment}{HTML}{566169}
\definecolor{CodeString}{HTML}{AF3A00}
\definecolor{CodeFunction}{HTML}{783C9B}
\AtBeginDocument{\hypersetup{linkcolor=InterfaceBlue,
  citecolor=InterfaceTeal,urlcolor=InterfaceBlue}}
\lstdefinestyle{xml}{language=XML,
  morekeywords={chain,joint,theta,link,transform,translation,rotationMatrix},
  morekeywords=[2]{type,dynamicJointType},
  keywordstyle=[2]\color{CodeFunction},morestring=[b]'}

\usepackage{colortbl}
\definecolor{TableMint}{HTML}{C9EEDA}
\definecolor{TableSalmon}{HTML}{FFD4C6}
\definecolor{TableHeader}{HTML}{D4E6F7}
\newcommand{\mintcell}[1]{\cellcolor{TableMint}{\bfseries\boldmath #1}}
\newcommand{\salmoncell}[1]{\cellcolor{TableSalmon}{\itshape #1}}

\AddToHook{env/table/begin}{  \setlength{\aboverulesep}{0pt}  \setlength{\belowrulesep}{0pt}  \setlength{\extrarowheight}{1.5pt}}

\begin{document}
\begin{frontmatter}
\title{\texttt{graspit-py}: A Python Interface to GraspIt! for Grasp Planning Under Pose and Friction Uncertainty}
\author[umd]{Clinton Enwerem\corref{corresponding}}
\ead{enwerem@umd.edu}
\author[umd]{John S. Baras}
\ead{baras@umd.edu}
\author[umd]{Calin Belta}
\ead{calin@umd.edu}
\cortext[corresponding]{Corresponding author: Clinton Enwerem. ORCID: \href{https://orcid.org/0000-0002-4239-2941}{0000-0002-4239-2941}.\ReportNote{}}
\affiliation[umd]{organization={Department of Electrical and Computer Engineering and Institute for Systems Research, University of Maryland},addressline={8223 Paint Branch Drive},city={College Park},state={MD},postcode={20742},country={USA}}
\begin{abstract}
In simulated multifingered grasps, pose error can change contacts formed during hand closure, while friction uncertainty changes the loads supported by those contacts. Existing C++ APIs and ROS~1 services provide programmatic GraspIt! access but require native bindings or a service transport layer for Python use and application-specific integration of contact geometry with uncertainty sampling. A Python workflow for these studies must connect native closure and contacts to caller-defined wrench scores, uncertainty samples, and command reconstruction. We present \texttt{graspit-py}, a Python interface to native GraspIt! operations that supplies this workflow. The interface provides native scene access, contact-wrench analysis, and configurable approach planning under pose and friction uncertainty. A grasp record associates the object-relative palm pose and closed independent coordinates with nominal contacts and quality scores evaluated under sampled pose and friction conditions. Our MuJoCo exporter derives serial hinge chains and affine joint couplings, including nonzero reference angles, and preserves individual collision components. Across 75 configurations of five hand models, maximum discrepancy between source and exported palm-frame vertex sets was approximately 1.3~\(\mu\)m against a 100~\(\mu\)m geometry tolerance. We isolated reference-angle and collision-component effects through controlled ablations and validated 1,920 directional task-wrench capacity calculations against independent Cartesian-force linear programs. In a common GraspIt! workload, median reset--close--score--state cycle latency through direct Python was 10.7\% lower than through persistent ROS~1 services. An executable export--plan--reconstruct workflow and grasp examples on six hand models demonstrate how the contact, score, and joint representations support numerical analysis and command reconstruction.

\end{abstract}
\end{frontmatter}
{\small\noindent Source Code: \url{https://github.com/coenwerem/graspit-py}\par}
\medskip
\section{Introduction}\label{introduction}

GraspIt! supports research on robotic hand design and grasp synthesis through a common environment for articulated hand modeling, object acquisition, and contact-wrench analysis \citep{miller_graspit_2004}. For comparisons of grasp generators or hand designs, the simulator supplies a consistent mechanical model while researchers specify the proposals and contact objectives. A learned proposal generator, for example, can supply palm poses (the palm's position and orientation) and finger configurations for closure, after which the resulting contacts can be evaluated using a custom wrench objective. A scene contains the loaded hand and object models, their placements, and hand configurations. Reusing GraspIt! in such studies requires programmatic control over its scenes and closure operations, together with access to the contact geometry. These capabilities allow its grasp mechanics to be incorporated into automated experiments alongside numerical optimization and data analysis.

Pose and friction uncertainty require different operations on a grasp: an error in the approach pose can move or remove contacts, requiring closure to be repeated, while a change in friction alters the admissible contact wrenches for each resulting contact set. Keeping contact positions and normals independent of the scoring routine allows researchers to apply their own wrench models and quality metrics to these closures. The Point Cloud Library (PCL) applies a similar separation to point-cloud processing, providing reusable data structures alongside modular filters and estimators \citep{rusu2011pcl}. The ROS~1 service interface and its Python \texttt{graspit\_commander} client expose simulator operations to external programs \citep{graspit_interface}, and a C++ API provides scene access and eigengrasp planning \citep{buehlerGraspItCppAPI2018}. We present \texttt{graspit-py}, which provides direct access to native GraspIt! closure and contact analysis within Python experiments. Its bindings connect the native engine to Python numerical routines without a ROS~1 service layer or application-specific native extensions. The interface combines this access with numerical mechanics and uncertainty utilities. Our planner reruns closure under pose perturbations, evaluates friction samples numerically, and retains the score samples so researchers can examine grasp sensitivity and select a statistical ranking appropriate to their study.

\begin{samepage}
Studies using planned grasps in task simulation also need to relate the hand used for grasp analysis to the robot model used for the task. Differences in articulation or collision geometry can change the contacts available to support an object, even when both models represent the same physical hand. Establishing this correspondence requires a joint-coordinate map and a comparison of the collision components attached to corresponding links. We support this connection by exporting hand and rigid-object geometry from a compiled MuJoCo model \citep{todorov2012mujoco} and retaining the joint correspondence needed to reconstruct a closed configuration in the source model. Figure~\ref{fig:motivation} illustrates the resulting research workflow: GraspIt! supplies closure contacts to Python for grasp synthesis and numerical evaluation, while the object-relative palm pose and joint map support reconstruction in MuJoCo. Researchers can thereby study alternative proposals and contact objectives using GraspIt!'s mechanical models and carry the selected configurations into a task simulator for subsequent analysis.

\end{samepage}

\input{figures/motivation.tex}

\subsection{Related Work}\label{related-work}

Grasp generation and grasp evaluation impose different requirements on a simulator interface. Eigengrasp planning searches a low-dimensional subspace of finger postures in GraspIt! \citep{ciocarlie_hand_2009}, while FRoGGeR optimizes fingertip placements using a differentiable quality measure \citep{li2023frogger}. Learned methods generate multifingered configurations or distributions over grasp poses from object geometry and partial views \citep{varley_generating_2015,wang2023dexgraspnet,lim_equigraspflow_2024,enweremEquiDexFlowContactGroundedSE32026}. Grasp taxonomies organize hand postures \citep{yang_grasp_2015}; image-conditioned and point-cloud methods such as GraspLDM, CenterGrasp, and REGNet provide geometric proposals \citep{barad_graspldm_2024,chisari_centergrasp_2024,zhao_regnet_2021}. The distinction between generating proposals and evaluating them is developed in data-driven grasp synthesis \citep{bohg2014data,mahler2017dex}. Our interface accepts arbitrary palm poses and independent coordinates. The provided planner evaluates combinations of approach direction, roll, preshape, and coordinates enabled for closure using native contact simulation. Consequently, proposal generation can be supplied by the caller while the resulting contacts remain available for numerical evaluation.

Contact-wrench analysis determines which external wrenches, or force--moment pairs, can be balanced by forces at the contacts. Analytic grasp measures distinguish force closure (the ability to resist wrenches in all directions), admissible contact forces, and the capacity to balance task loads \citep{mason_robot_1985,bicchi_robotic_2000,roaGraspQualityMeasures2015,zheng_limiting_2006,zheng_computing_2012}. Uncertain friction and displaced contacts change the wrench hull and its force-closure properties \citep{zheng_coping_2005}. Pose-based re-ranking evaluates grasp sensitivity to object-pose error \citep{weisz_pose_2012}, PONG bounds force-closure probability under uncertain surface normals \citep{li2023pong}, and intrinsic-robustness results relate friction-cone uncertainty to Ferrari--Canny and min-weight quality \citep{li2024intrinsic}. For force-closure grasps under a specified force budget and torque scaling, Ferrari--Canny quality is the radius of the largest origin-centered Euclidean ball contained in the normalized wrench hull. Grasp'D differentiates through simulated contact \citep{turpin_graspd_2022-1}, and elastic-fingertip stability can be evaluated through potential energy \citep{tsuji_grasp_2014}. These quantities describe different physical properties. We expose native contact geometry separately from the numerical wrench calculation so the friction model, force budget, and task loads used in a score can be specified explicitly.

Uncertainty-aware planning also depends on how uncertain quantities are represented and updated. Partially observable decision processes distinguish state uncertainty from the observations available to a planner \citep{lauriPartiallyObservableMarkov2023}. Encoder feedback and particle filtering can reduce pose uncertainty during planar grasp sequences \citep{zhouProbabilisticPlanningFramework2017}, and hypothesis-based belief planning uses contact observations in reach-to-grasp trajectories with non-Gaussian pose uncertainty \citep{zitoHypothesisbasedBeliefPlanning2019}. Variational belief models represent uncertainty over contact parameters \citep{enweremVariationalNeuralParameterizations2026a}, while risk-sensitive incremental sampling applies a risk criterion to stochastic shortest-path costs \citep{enwerem2024robust}. FIRMGrasp evaluates a force-closure margin at a friction coefficient computed from the adverse tail of its distribution \citep{enweremFIRMRobustGraspSynthesis2026}. In the present planner, the decision is an approach command and the sampled quantity is the wrench quality after closure. Pose perturbations trigger new closures, and friction values are evaluated for each resulting contact set before the lower-tail statistic is formed.

Transferring such commands requires agreement between the planning and execution models. Manipulator kinematics determine link poses from joint coordinates \citep{siciliano2010robotics}; a change in hinge axes or reference frames changes fingertip positions even when coordinate values agree. The ROS-based urdf2graspit converter maps URDF kinematics and meshes into GraspIt! descriptions \citep{buehler2018urdf2graspit}. Our exporter starts from a compiled MuJoCo model and retains the affine joint correspondence and individual collision components used by the selected geometric representation. The resulting joint and link maps make the correspondence measurable at specified configurations and support reconstruction of a planned object-relative grasp in MuJoCo.

Contact geometry is also distinct from the feedback and dynamics governing acquisition and retention. Closed-loop depth-based grasping and learned action values update commands using observations or trial outcomes \citep{morrison2018closing,kalashnikov2018scalable}. Tactile feedback can guide grasp acquisition and regrasping \citep{dang_blind_2011,calandra2018more}, regulate forces or stabilize individual fingers \citep{deng_grasping_2020,veiga_grip_2020}, and supply stability predictors for transfer \citep{si_grasp_2022}. Dynamic grasp evaluation accounts for object motion and contact changes during repeated acquisitions and lifts under pose uncertainty \citep{kim2013physically}; compliant grasp synthesis additionally models finger impedance under shape uncertainty \citep{chen2024springgrasp}. These methods explain why the closing motion and contact response matter alongside a nominal wrench score. In assembly, compliant contact-rich motion \citep{chen2025robust} and redundant-arm motion--force control \citep{fan_data-driven_2021} act on the grasped object, while simulation systems such as Orbit supply robot-learning environments \citep{mittal2023orbit}. The wrapper supplies contact geometry and reconstructed commands to these external planning, simulation, and control components.

\subsection{Contributions}\label{contributions}
We contribute a Python interface for acquiring native GraspIt! closures and evaluating their contact wrenches under pose and friction uncertainty. The scene, hand, object, contact, and link queries are complemented by numerical kinematics and wrench-space utilities (Section~\ref{software-architecture}). For MuJoCo hands with serial revolute finger chains and hinges coupled by fixed ratios and offsets, we derive the affine joint map, including reference-angle offsets, and the rigid-frame transformation needed to preserve individual collision components (Section~\ref{mujoco-model-export}). Grasp records associate the accepted palm pose and closed independent coordinates with nominal contacts and pose--friction score arrays, making the scoring inputs and reconstruction data explicit (Sections~\ref{contact-based-planning} and~\ref{sec:runnable-workflow}). We assessed the interface through complementary geometric, numerical, and runtime comparisons: link-by-link correspondence and controlled exporter ablations, agreement with an independent Cartesian-force formulation, and runtime and resident-memory scaling over multi-hand closure and uncertainty sampling (Section~\ref{verification}).

\section{Software Architecture}\label{software-architecture}

\begin{figure}[!htbp]
\centering
\begin{tikzpicture}[x=1mm,y=1mm,
 box/.style={draw=black!55,rounded corners=1.5pt,align=center,font=\small,inner sep=3pt},
 native/.style={box,fill=blue!7},numeric/.style={box,fill=TableMint},
 record/.style={box,fill=TableHeader},client/.style={box,fill=TableSalmon},
 flow/.style={-{Stealth[length=1.7mm]},semithick}]
\node[record,text width=29mm] (source) at (0,38) {Compiled MuJoCo\\hand and object};
\node[numeric,text width=34mm] (export) at (47,38) {Model Conversion\\XML, OFF, joint map};
\node[native,text width=35mm] (world) at (99,38) {Native \texttt{World}\\scene, joints, closure};
\node[record,text width=29mm] (config) at (0,9) {Python / CLI\\approaches, constraints,\\loads, pose / friction law};
\node[native,text width=34mm] (planner) at (47,9) {\texttt{ContactPlanner}\\approach and closure\\pose samples / filters};
\node[numeric,text width=35mm] (quality) at (99,9) {Numerical Analysis\\wrench matrix, \(\varepsilon\),\\min-weight, \(Q\)};
\node[record,text width=55mm] (records) at (48,-21) {Grasp Records\\object-relative palm pose, DOFs, nominal contacts, score arrays};
\node[client,text width=35mm] (external) at (108,-21) {Application Clients\\arm planning, simulation,\\feedback control};
\draw[flow] (source)--(export);
\draw[flow] (export)--node[above,font=\scriptsize]{models}(world);
\draw[flow] (config)--(planner);
\draw[flow] (planner)--(world);
\draw[flow] (world)--node[right,font=\scriptsize]{contacts / link queries}(quality);
\draw[flow] (planner)--(records);
\draw[flow] (quality.south)--++(0,-9)-|(records.north east);
\draw[flow] (records)--(external);
\draw[dashed,black!55] (84,-37)--(84,-8);
\node[font=\scriptsize,align=left,anchor=west] at (-16,-39) {Model conversion and numerical analysis run without native bindings.\\Scene handles share the native world within each worker process.};
\end{tikzpicture}
\caption{\textbf{Wrapper Architecture.} Model conversion supplies hand and object geometry and the joint map to native GraspIt!. Candidate commands and their pose perturbations undergo approach and closure, producing contacts for numerical wrench analysis. Grasp records retain the accepted object-relative palm pose, closed independent coordinates, nominal contacts, and sampled scores for subsequent selection or reconstruction.}
\label{fig:architecture}
\end{figure}
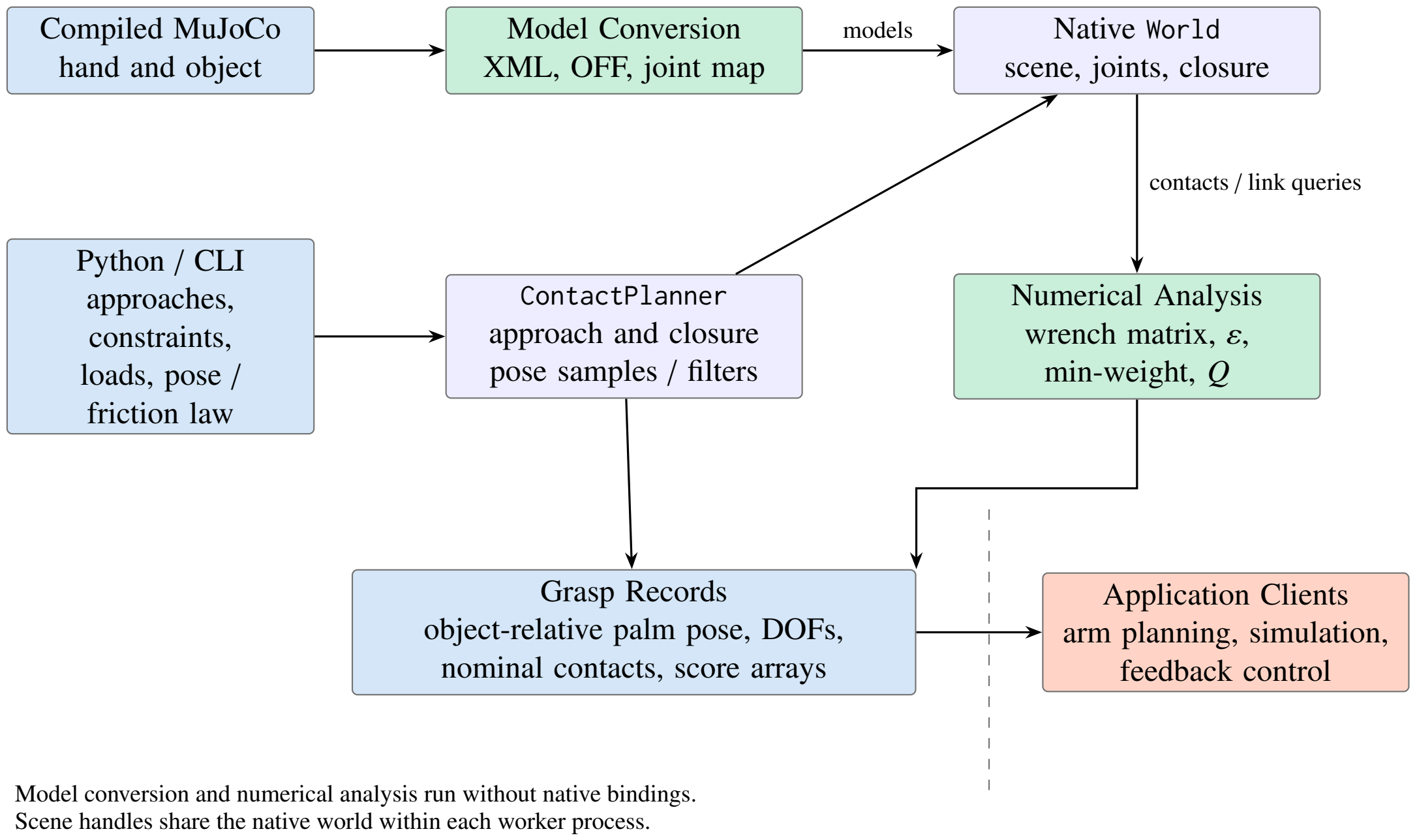

We separate native scene operations from array-based analysis so researchers can use the same closure model with different candidate generators and numerical objectives. Native GraspIt! supplies hand motion, collision queries, and contact geometry. The planning module combines these operations with pose perturbations and friction-dependent scoring, while model conversion supplies the geometry and source joint correspondence needed for reconstruction. Figure~\ref{fig:architecture} shows these data dependencies, and Table~\ref{tab:interfaces} identifies the public modules. The interfaces below make the coordinate, contact, and score representations explicit.

\begin{table}[t]
\centering
\caption{\scshape Package Interfaces}\label{tab:interfaces}
\small
\begin{tabularx}{\linewidth}{>{\raggedright\arraybackslash}X>{\raggedright\arraybackslash}X>{\raggedright\arraybackslash}X}
\toprule
\rowcolor{TableHeader}
Public module & Input & Output \\
\midrule
\texttt{core.bindings} & Robot and object models, hand pose, joint coordinates & Closure, object-frame contacts, native quality, collision and link queries \\
\texttt{convert.}\newline\texttt{mjcf\_hand} & Compiled MuJoCo model and selected palm or object subtree & GraspIt! XML and OFF meshes, with an affine source joint map for hand export \\
\texttt{planning} & Palm-frame geometry, approaches, preshapes, closing subset, constraints, and uncertainty parameters & Ranked records with object-relative palm poses, independent coordinates, contacts, and scores \\
\texttt{kinematics} & GraspIt! robot XML and independent coordinates & Link forward kinematics \\
\texttt{grasp.quality}, \texttt{risk} & Contacts and load directions, or weighted score samples & Wrench-space metrics, task load multiplier, and tail averages \\
\texttt{utils.geometry}, \texttt{utils.urdf} & Geometric data or a rigid URDF object with optional dependencies & Transforms, distances, or OFF/XML object models \\
\bottomrule
\end{tabularx}
\end{table}

\subsection{Scene State and Hand Motion}\label{sec:scene-state}

Before running closure experiments, the caller initializes a GraspIt! session and loads the required hand and object models. The native extension must be importable; a missing extension raises \texttt{ImportError} when the bindings are imported. The caller invokes the zero-argument \texttt{initialize()}, which returns \texttt{None}, then constructs a \texttt{World} handle and calls \texttt{World.init()} before loading robot, body, or world descriptions. The environment variable \texttt{GRASPIT} identifies the directory containing GraspIt!'s hand models, object descriptions, and world files, and Qt uses \texttt{DISPLAY} and \texttt{QT\_QPA\_PLATFORM}; when creating the Qt application, initialization selects the \texttt{offscreen} platform if both Qt variables are unset. Native initialization runs once per Python process, so repeated calls reuse the initialized core. Initialized \texttt{World} handles share the main GraspIt! scene; independent scenes require separate worker processes.

Hand posture is \(u\in\mathbb{R}^m\), the vector of \(m\) independent coordinates in radians, ordered by the loaded model. One entry of \(u\) can command several hinges, as when the joints nearest the palm and fingertip close together through the coupling in Section~\ref{mujoco-model-export}. A palm pose uses the quaternion and world-frame millimeter translation conventions in Table~\ref{tab:api-conventions}. Direct pose and posture setters establish the requested state. Posture assignment applies native coordinate-limit handling without collision checking. The native scene check \texttt{World.no\_collision()} returns \texttt{True} precisely when no pair of loaded world bodies interpenetrates, subject to native collision filtering.

Controlled closure instead uses \texttt{move\_current\_hand\_dof\_to\_contacts} with a target coordinate vector and optional per-coordinate step sizes. Native small-step collision checking governs this motion. Its traced variant returns a Boolean reporting at least one successful native movement step and records the independent coordinates and collision count at each step signal. These records expose the progression of a closing posture without assigning time or forces to its steps. The native \texttt{auto\_grasp} operation uses the hand's configured default velocities; setting selected velocities to zero holds their coordinates at the preshape. The planner uses this mechanism for closing subsets and restores the original velocities after closure. Table~\ref{tab:callable-contracts} summarizes the corresponding input and output contracts.

\subsection{Contact Geometry and Force Queries}\label{sec:contact-queries}

After changing the pose or posture and closing the hand, the caller refreshes a \texttt{Grasp} with \texttt{compute()}. Its \texttt{object\_contact\_geometry()} query returns object-side contact positions \(p_i\) in millimeters and inward normals \(n_i\), both in the object frame. This common representation supplies numerical wrench analysis together with the object-frame center of mass \(c\). Matrix construction takes a friction coefficient and torque scale \(\rho\), producing the wrench columns \(W_m\) defined in Section~\ref{contact-based-planning}. The array-based modules operate independently of native scene state, using NumPy and SciPy \citep{harris2020numpy,virtanen2020scipy}. Native quality \(\varepsilon_{\mathrm{GI}}\), recomputed Ferrari--Canny quality \(\varepsilon\), and the min-weight optimum retain their respective definitions; the min-weight routine also returns a rank-qualified force-closure Boolean.

Quasistatic force queries require a separate native solve. The caller checks \texttt{compute\_quasistatic\_wrenches()} and the availability flag, or uses optional normal-force values to retain missing results explicitly. Calling \texttt{compute()} refreshes the contacts and clears the cached quasistatic solution. Force queries for the refreshed contact set trigger a new solve when no solve has yet been attempted. Native contact wrenches have force components preceding moment components and retain their native units and contact-frame axes. For a solved compressive contact, let \(f_{n,i}>0\) and \(f_{t,i}\) denote its normal and tangential force components in this frame. The required friction coefficient is
\begin{equation}
 \mu_{i,\mathrm{req}}=\frac{\|f_{t,i}\|_2}{|f_{n,i}|}.
 \label{eq:friction-demand}
\end{equation}
Comparing \(\mu_{i,\mathrm{req}}\) with the contact coefficient \(\mu_i\) tests the friction demand of the solved forces. For solved compressive contacts with normal force and friction coefficient above their numerical zero thresholds, the returned friction-demand indicator is \(\min\{\mu_{i,\mathrm{req}}/\mu_i,1\}\). Small values indicate low tangential force relative to the Coulomb limit, while unity marks a demand at or above the limit.

\begin{table}[t]
\centering
\caption{\scshape Coordinate and Data Conventions}\label{tab:api-conventions}
\small
\begin{tabularx}{\linewidth}{>{\raggedright\arraybackslash}X>{\raggedright\arraybackslash}X>{\raggedright\arraybackslash}X}
\toprule
\rowcolor{TableHeader}
Representation & Frame and Units & Interface Convention \\
\midrule
Native palm pose & World frame; translation in mm & Quaternion \((w,x,y,z)\); pose query includes a nested \texttt{matrix\_row\_major} with four rows and four columns \\
Grasp-record palm pose & Object frame; translation in mm & \texttt{T\_object\_hand\_mm} maps palm coordinates into object coordinates \\
Independent coordinates & Loaded hand's native order; radians & Shared drivers can control several hinges \\
Object-side contacts & Object frame; positions in mm & \texttt{position\_mm} and inward force \texttt{normal} \\
Native link queries & World frame; vertices in mm & Poses identify \texttt{chain}, \texttt{link}, and \texttt{name} \\
XML forward kinematics & Palm frame; XML distances typically mm & \texttt{link\_poses} returns homogeneous transforms keyed by chain and link \\
Geometry utilities & Caller-supplied frame and length unit & Quaternions \((x,y,z,w)\); mesh distances are unsigned \\
\bottomrule
\end{tabularx}
\end{table}

\subsection{Geometry and Kinematic Queries}\label{sec:geometry-queries}

Frame composition connects native geometry, exported kinematics, and source reconstruction. Let \(\{W\}\), \(\{O\}\), and \(\{H\}\) denote the source world, object, and palm frames, and let \(\{L\}\) denote a link frame. With translations expressed in a common length unit,
\begin{equation}
 {}^{W}T_L={}^{W}T_O\,{}^{O}T_H\,{}^{H}T_L(u).
 \label{eq:api-link-composition}
\end{equation}
Native link queries return poses and vertices in the native scene's world frame. Pure forward kinematics reads the exported Denavit--Hartenberg description and evaluates palm-frame link poses from independent coordinates \(u\). For comparison, the caller expresses native and source link poses relative to their corresponding object frames by applying each scene's inverse object pose. The common object-frame correspondence permits comparison with \({}^{O}T_H{}^{H}T_L(u)\) using the chain and link identities. Geometry utilities transform points and compute unsigned point--triangle or point--mesh distances in the input length unit. Native poses use scalar-first quaternions \((w,x,y,z)\), while geometry utilities use scalar-last quaternions \((x,y,z,w)\) (Table~\ref{tab:api-conventions}). The caller reorders components when transferring an orientation between these interfaces. The export manifest supplies source joint names, driver indices, and affine coefficients, while \texttt{dofs\_from\_qpos} extracts the independent coordinates by joint name and position address.

The optional rigid-object utility \texttt{urdf\_to\_graspit\_object\_xml} combines URDF visual meshes into OFF geometry and writes a GraspIt! object XML description. The caller supplies the length scale and mass in grams to prepare the rigid object for native loading and closure studies. This visual-mesh preparation is distinct from the compiled-MuJoCo hand and object exporters in Section~\ref{mujoco-model-export}, which use the source model's collision geometry.

\begin{table}[t]
\centering
\caption{\scshape Selected Callable Contracts}\label{tab:callable-contracts}
\small
\begin{tabularx}{\linewidth}{>{\raggedright\arraybackslash}X>{\raggedright\arraybackslash}X>{\raggedright\arraybackslash}X}
\toprule
\rowcolor{TableHeader}
Callable Family & Caller Supplies & Result or State Change \\
\midrule
\texttt{World} loading and setters & Model paths, world-frame palm pose, independent coordinates & Shared native scene state; loading success Booleans \\
\texttt{move\_current\_hand\_}\newline\texttt{dof\_to\_contacts} and traced variant & Desired coordinates and optional step sizes & Collision-aware coordinate motion; optional coordinate/collision-count trace \\
\texttt{Grasp.compute}, \texttt{object\_contact\_geometry} & Loaded hand, selected object, current closure & Refreshed grasp and object-frame contact dictionaries \\
Quasistatic solve and force queries & Current grasp & Solve status, availability, contact-frame wrenches and normal-force diagnostics \\
\texttt{build\_grasp\_matrix}, \texttt{min\_weight\_force\_closure} & Common-frame contacts, center, friction, torque scale & Wrench matrix; min-weight optimum and force-closure predicate \\
\texttt{load\_chains}, \texttt{link\_poses} & Robot XML and independent coordinates & Palm-frame link transforms keyed by chain and link \\
\texttt{lower\_tail\_mean} & Score values, tail mass, optional weights & Weighted lower-tail quality average \\
\texttt{GraspCandidate.to\_dict} & Existing grasp-record object & Palm placement, closed coordinates, nominal contacts, sampled scores \\
\bottomrule
\end{tabularx}
\end{table}

\subsection{Approach Planning and Grasp Records}\label{sec:grasp-records}

The planner enumerates combinations of approach direction, roll, preshape, and closing subset. Callers can supply explicit approaches or use Fibonacci directions and equally spaced rolls; each closing subset is a sequence of integer indices into the independent-coordinate vector. The \texttt{closing\_sets} configuration enumerates these subsets as alternative grasp commands. The pure \texttt{palm\_pose\_for\_approach} function constructs the planner's standoff transform for use by an external reachability check. Each native approach resets the preshape and advances through discrete collision checks before closing at the accepted placement. The planner retains the object's construction-time pose for these calculations. Minimum-contact and allowed-contact-region tests apply to nominal and perturbed closures. Support-plane and keep-out-box tests inspect final nominal hand vertices, so their scope is the sampled vertices at this placement. Task rows provide object-frame application points and balancing-force directions, normalized individually before scoring. Section~\ref{contact-based-planning} develops the approach, load, and uncertainty calculations.

The dictionary returned by \texttt{GraspCandidate.to\_dict()} retains the attempted command and its nominal closure (Table~\ref{tab:grasp-record-schema}). NumPy arrays become Python lists. For \(n_\xi\) perturbed approaches and \(n_\mu>0\) friction nodes, the score arrays have shape \((n_\xi+1,n_\mu)\) and represent the empirical law in Equation~\ref{eq:empirical-pose-friction-law}. With zero-based indexing, row zero is nominal, and nominal scalar fields use column \(\lfloor n_\mu/2\rfloor\), the upper middle column when \(n_\mu\) is even. All scores are dimensionless under the documented normalization, and \(\beta\) denotes the lower-tail mass. The task fields are \texttt{None} without task loads. A command-dependent random stream preserves its samples under a fixed seed and configuration, independently of enumeration order. The caller retains the search configuration with the record to identify the friction grid, pose distribution, and aggregation parameters.

\begin{table}[t]
\centering
\caption{\scshape Serialized Grasp Record Schema}\label{tab:grasp-record-schema}
\small
\begin{tabularx}{\linewidth}{>{\raggedright\arraybackslash}X>{\raggedright\arraybackslash}X>{\raggedright\arraybackslash}X}
\toprule
\rowcolor{TableHeader}
Key & Serialized Type, Shape, and Units & Meaning \\
\midrule
\texttt{T\_object\_hand\_mm} & Nested float lists, \(4\times4\); translation in mm & Accepted nominal palm-to-object transform. \\
\texttt{dofs\_rad} & Float list, length \(m\); rad & Closed independent coordinates. \\
\texttt{contacts} & List of dictionaries with length-three float lists & Nominal object-frame \texttt{position\_mm} (mm) and inward unit \texttt{normal}. \\
\texttt{native\_epsilon} & Float & Native GraspIt! quality of the nominal closure. \\
\texttt{nominal\_epsilon} & Float & Nominal-row recomputed quality at the middle friction node. \\
\texttt{epsilon\_samples} & Nested float lists, \((n_\xi+1)\times n_\mu\) & Recomputed Ferrari--Canny qualities over poses and friction nodes. \\
\texttt{score} & Float & Ranking score: \texttt{epsilon\_cvar} without task loads; lower-tail mean of \texttt{task\_samples} with mass \(\beta\) otherwise. \\
\texttt{expected\_epsilon} & Float & Mean over every \texttt{epsilon\_samples} cell. \\
\texttt{approach\_dir} & Float list, length three & Object-frame unit approach direction. \\
\texttt{roll\_deg} & Float; degrees & Attempted approach roll. \\
\texttt{preshape} & Float list, length \(m\); rad & Attempted independent coordinates before closure. \\
\texttt{closing} & Integer list or \texttt{None} & Selected zero-based indices from \(0\) to \(m-1\); \texttt{None} enables all coordinates. \\
\texttt{epsilon\_cvar} & Float & Lower-tail mean of \texttt{epsilon\_samples} with mass \(\beta\). \\
\texttt{task\_samples} & Nested float lists, \((n_\xi+1)\times n_\mu\), or \texttt{None} & Task-capacity samples over poses and friction nodes. \\
\texttt{nominal\_task} & Float or \texttt{None} & Nominal-row task capacity at the middle friction node. \\
\texttt{expected\_task} & Float or \texttt{None} & Mean over every \texttt{task\_samples} cell. \\
\bottomrule
\end{tabularx}
\end{table}

\subsection{MuJoCo Model Export}\label{mujoco-model-export}

Native closure requires GraspIt! descriptions of the hand and object, including their geometry and articulation. We call the MuJoCo model supplied for conversion the \emph{source model}; its in-memory \texttt{MjModel} representation is the \emph{compiled model}. The \emph{MuJoCo source scene} includes the hand, object, and surrounding scene geometry. The hand exporter takes this compiled model and the name of the palm body. A link is a rigid segment of the hand. A finger forms a serial chain when its moving links connect sequentially without branching. Each supported moving link has one revolute joint, so its relative motion is rotation about a fixed hinge axis \citep{siciliano2010robotics}. The hand consists of several such chains attached to the palm. A jointless body is rigidly attached to its nearest exported ancestor, and its collision geometry is expressed in the ancestor's link frame.

Several hinges can share one independent command. Write \(u\in\mathbb{R}^{m}\) for the independent coordinates and \(q\in\mathbb{R}^{n}\) for the hinge angles in radians. Joint \(i\) is parameterized by
\begin{equation}
 q_i=r_i u_{j(i)}+b_i,
 \qquad q_i^{\min}\leq q_i\leq q_i^{\max}.
 \label{eq:joint-coupling}
\end{equation}
Here \(j(i)\) identifies the driver, \(r_i\) is a dimensionless ratio, and \(b_i\) is an angular offset. An independent hinge has \(r_i=1\) and \(b_i=0\). A MuJoCo equality couples a driven hinge \(i\) to an independent hinge \(j\) through the displacements from their reference angles \citep{mujocoxmlreference}. For an affine equality, the conversion to absolute angles is
\begin{equation}
 \begin{aligned}
 q_i-q_i^0 &= c_0+c_1(q_j-q_j^0), \\
 r_i &= c_1, \qquad b_i=q_i^0+c_0-c_1q_j^0.
 \end{aligned}
 \label{eq:reference-coupling}
\end{equation}
The exporter reads each reference angle at its joint's position address in the compiled model and writes the resulting absolute-coordinate mapping to the manifest. This mapping also determines the angle expression and limits of the GraspIt! joint. The supported model class uses direct affine couplings to independent hinges.

The manifest expresses the scalar coupling relation as a global coordinate map. Let \(q\in\mathbb R^n\) collect source hinge angles in a fixed joint-name order, and let \(u\in\mathbb R^m\) contain independent coordinates in the manifest's \texttt{dof\_joints} order. Then
\begin{equation}
 q=Au+b,\qquad
 A_{ij}=\begin{cases}r_i,&j=d_i,\\0,&j\ne d_i,\end{cases}
 \label{eq:global-coupling-map}
\end{equation}
where \(d_i\) selects the driver of hinge \(i\), \(r_i\) is its ratio, and \(b_i\) is the reference-corrected offset in \eqref{eq:reference-coupling}. Angles and offsets are in radians; \(A\in\mathbb R^{n\times m}\) is dimensionless. Independent hinge rows contain the corresponding rows of \(I_m\) with zero offsets, so \(\operatorname{rank}A=m\), including when a follower has zero ratio. Joint names, driver indices, ratios, offsets, and source intervals in the manifest provide the data needed to construct the map.

Source limits constrain the independent coordinates through the same affine map. For a hinge interval \([q_i^-,q_i^+]\) and \(r_i\ne0\), its admissible driver interval is
\begin{equation}
 \begin{aligned}
 L_i&=\min\left\{\frac{q_i^--b_i}{r_i},
                    \frac{q_i^+-b_i}{r_i}\right\},\\
 U_i&=\max\left\{\frac{q_i^--b_i}{r_i},
                    \frac{q_i^+-b_i}{r_i}\right\}.
 \end{aligned}
 \label{eq:coupled-hinge-interval}
\end{equation}
The endpoint ordering handles negative ratios. A zero-ratio hinge instead requires \(q_i^-\le b_i\le q_i^+\). A feasible constant hinge adds no restriction on its driver, while an infeasible constant hinge makes the complete coordinate domain empty.

For the supported direct couplings, each nonconstant hinge constrains one independent coordinate. When all zero-ratio hinges satisfy \(q_i^-\le b_i\le q_i^+\), intersecting all intervals assigned to driver \(j\), including its own hinge interval, gives
\begin{equation}
 \begin{aligned}
 \mathcal U&=\{u:q^-\le Au+b\le q^+\}
       =\prod_{j=1}^{m}[\underline u_j,\overline u_j],\\
 \underline u_j&=\max_{i:d_i=j,\ r_i\ne0}L_i,
 &\overline u_j&=\min_{i:d_i=j,\ r_i\ne0}U_i.
 \end{aligned}
 \label{eq:independent-coordinate-domain}
\end{equation}
The box is feasible when all constant hinges satisfy their limits and \(\underline u_j\le\overline u_j\) for every driver. The example utility \texttt{coupled\_dof\_bounds} computes the intersections and rejects infeasible or unbounded results. It uses exported source intervals; a source hinge without limits receives \([-\pi,\pi]\) during export.

The \texttt{dofs\_from\_qpos} helper recovers independent coordinates by selecting the corresponding hinge angles from a source posture. Let \(E\in\mathbb R^{m\times n}\) be the corresponding selector matrix. Its identity rows imply \(EA=I_m\) and \(Eb=0\); hence
\begin{equation}
 u=Eq,\qquad A(Eq)+b=q
 \quad\text{for }q\in\{Au+b:u\in\mathbb R^m\}.
 \label{eq:coupling-reconstruction}
\end{equation}
Selection uses each source joint's name and position address. Exact reconstruction holds on the affine coupling subspace. For an arbitrary source posture, a caller can compute \(q-A(Eq)-b\) to diagnose disagreement of driven hinges with the exported relation. Selection itself neither projects the posture onto the coupling subspace nor tests its residual. The manifest's joint identities preserve the correspondence needed to interpret every component.

We construct Denavit--Hartenberg frames from the hinge axes in the model reference configuration \citep{siciliano2010robotics,miller_graspit_2004}. Let \(B\) map the base frame of a finger to the palm frame. For link \(i\) in the chain, the palm-frame pose is
\begin{equation}
 {}^{H}T_i(u)=B\prod_{k=1}^{i}
 \bigl[R_z(\theta_k(u))\,T_z(d_k)\,T_x(a_k)\,R_x(\alpha_k)\bigr],
 \qquad
 \theta_k(u)=\theta_k^0+q_k-q_k^0.
 \label{eq:dh-chain}
\end{equation}
The product follows the chain from the palm outward. The operators \(R_x,R_z\) denote rotations and \(T_x,T_z\) translations. The constant \(\theta_k^0\) is the reference Denavit--Hartenberg angle, \(d_k\) is the distance along the preceding hinge axis, \(a_k\) is the common-normal distance, and \(\alpha_k\) is the twist about the common normal. Both distances are in millimeters. The exporter writes angular constants in degrees in the robot XML and re-expresses collision meshes in the computed link frames. For successive unit hinge directions \(\hat z_k,\hat z_{k+1}\) with \(\|\hat z_k\times\hat z_{k+1}\|_2<10^{-2}\), the parallel-axis construction sets \(d_k=0\) to avoid distant common-normal endpoints. For nearly parallel axes this construction can approximate the original geometry, so the resulting correspondence is evaluated numerically in Section~\ref{geometry-in-native-graspit}.

The output includes robot XML, link meshes, and a manifest identifying corresponding links and joint names. We distinguish the source mesh used for visualization from its collision representation. MuJoCo attaches geometric elements, called \emph{geoms}, to rigid bodies for visualization and collision detection. For a standard mesh geom participating in collision detection, MuJoCo uses a convex hull, even when the source surface has concavities \citep{mujococollisionreference}. Copying source triangles preserves their coordinates but can change the surface on which closure stops. With \texttt{mesh\_collision=}\allowbreak\texttt{'convex\_hull'}, the exporter writes the hull vertices and faces stored in the compiled model. Each geom remains an individual convex component. Thus a body represented by several convex geoms retains their union; a hull over all their vertices would fill the spaces between components. The default \texttt{'source'} mode retains the source triangle surfaces. The exporter converts supported analytic shapes (boxes, spheres, ellipsoids, capsules, and cylinders) into triangle meshes in both modes. The object exporter uses the named body frame and carries over mass and center of mass, providing the object frame for scoring and command reconstruction.

To state the geometric correspondence, let \(K_g\subset\mathbb R^3\) be the compiled hull of geom \(g\) in its geom frame, and let \({}^{B}T_g\) map this frame to the attached body frame \(\{B\}\). Let \(\{L\}\) be its exported Denavit--Hartenberg link frame, with constant transform \(C_B={}^{B}T_L\). We express every geometric set and transform translation in millimeters. We write the component in the exported frame as
\begin{equation}
 \widehat K_g^{L}=C_B^{-1}\,{}^{B}T_gK_g.
 \label{eq:collision-frame}
\end{equation}
A rigid transform acts on every point of the set. If the corresponding link kinematics satisfy \({}^{H}T_L(u)={}^{H}T_B(u)C_B\), then
\begin{equation}
 {}^{H}T_L(u)\widehat K_g^{L}
 ={}^{H}T_B(u)\,{}^{B}T_gK_g.
 \label{eq:collision-correspondence}
\end{equation}
Consequently, corresponding components and their unions occupy equal sets in the palm frame. Rigid transformations also commute with convexification, so converting the frames does not change the hull. This identity applies to configurations satisfying the affine joint map in \eqref{eq:joint-coupling}. For a tessellated primitive with Hausdorff approximation error \(\delta_g\), a rigid transform preserves the error, and the error of a finite union of paired components is at most \(\max_g\delta_g\). In the geometry evaluation, we compared corresponding palm-frame vertices in each export representation and checked collision distances and component connectivity. A collision-pair exclusion disables collision checking between specified bodies, such as adjacent finger links. These settings must agree when closure behavior is compared across simulators.

\noindent\begin{minipage}{\linewidth}
\begin{lstlisting}[language=Python,caption={Export a Hand and a Rigid Object from a Compiled Scene},label={lst:export}]
import mujoco
from graspit_py.convert.mjcf_hand import (
    export_body, export_hand)

model = mujoco.MjModel.from_xml_path("scene.xml")
hand = export_hand(model, "palm", "out/hand", name="hand")
object_xml = export_body(model, "key", "out/key")
\end{lstlisting}
\end{minipage}

The input body names \texttt{palm} and \texttt{key} identify the hand root and rigid object. The generated robot and object paths are inputs to the planner, while the manifest supplies the joint correspondence for source-model reconstruction. The executable workflow in Section~\ref{sec:runnable-workflow} uses these exports in a complete planning instance.

Listing~\ref{lst:l6-xml} shows the index-finger coupling in the exported \texttt{l6\_right.xml}. Both Denavit--Hartenberg angle expressions use the independent coordinate named \texttt{d2} in the XML. The proximal angle adds a constant offset in degrees, and the distal angle scales the coordinate by 1.125676. We export twelve link meshes and six independent coordinates for the RealHand L6.

\noindent\begin{minipage}{\linewidth}
\begin{lstlisting}[style=xml,caption={Exported GraspIt! XML Excerpt for the RealHand L6 Index Finger},label={lst:l6-xml}]
<chain>
  <joint type='Revolute'><theta>d2+-88.085737930</theta></joint>
  <joint type='Revolute'><theta>d2*1.125676+0</theta></joint>
  <link dynamicJointType='Revolute'>rh_index_proximal.xml</link>
  <link dynamicJointType='Revolute'>rh_index_distal.xml</link>
</chain>
\end{lstlisting}
\end{minipage}

\begin{figure}[!htbp]
\centering
\begin{minipage}[b]{0.49\linewidth}\centering
{\setlength{\fboxrule}{1pt}\setlength{\fboxsep}{0pt}\fbox{\includegraphics[width=\dimexpr\linewidth-2\fboxrule\relax]{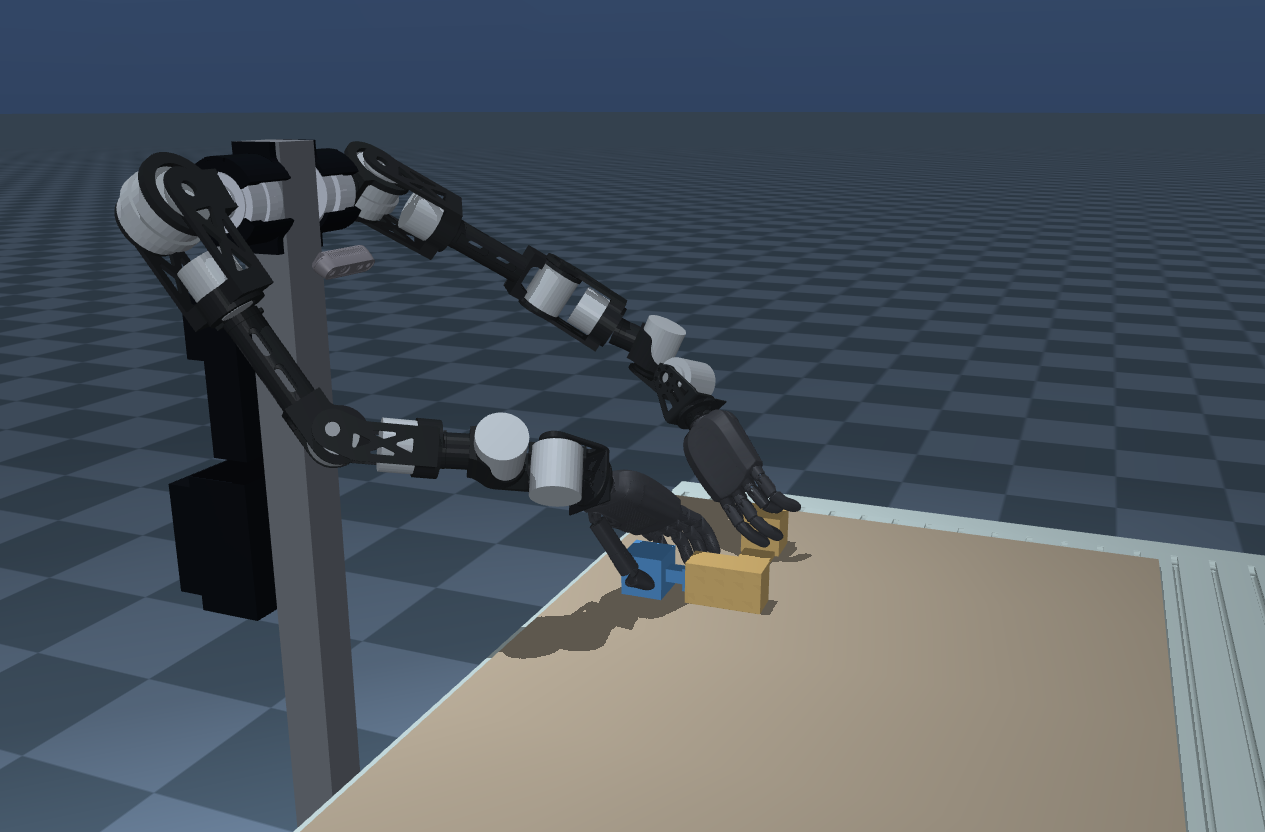}}}\par
\smallskip\small (a)
\end{minipage}\hfill
\begin{minipage}[b]{0.49\linewidth}\centering
\input{figures/export-grasp-frames.tex}\par
\smallskip\small (b)
\end{minipage}
\caption{\textbf{MuJoCo Source Scene and Exported Hand.} (a) MuJoCo scene with the source RealHand L6 models in an insertion state. (b) GraspIt! export of the right hand at the planned key grasp, with link poses queried in GraspIt!. The axes mark the palm frame \(\{H\}\) and key frame \(\{O\}\). The callouts identify the five digit chains.}
\label{fig:export}
\end{figure}

We transfer the object-relative palm pose to the MuJoCo source scene and reconstruct the finger configuration using the joint map (Fig.~\ref{fig:export}).

\subsection{Contact Analysis and Uncertainty-Aware Planning}\label{contact-based-planning}

The planner evaluates grasp commands by simulating approach and closure, then scoring the resulting contacts under pose and friction uncertainty. For each approach command, the hand advances from its standoff pose toward the object and closes the selected independent joints from the specified preshape. Task constraints restrict where fingers may touch the object and which surrounding regions the hand may occupy. A minimum contact count rejects closures with too few recorded contacts, while an allowed contact region can preserve access to a tool's working end. A support plane specifies the allowed side of a supporting surface for the nominal hand vertices, and keep-out boxes represent fixture regions excluded from those vertices. For a box centered at \(k\) with half-widths \(h\), a vertex \(v\) lies in its excluded interior when \(|v_j-k_j|<h_j\) for every coordinate \(j\); a boundary point does not satisfy this strict interior predicate. The contact-count and contact-region filters also apply to perturbed closures. The accepted palm pose, closed coordinates, nominal contacts, and sampled scores form the grasp record defined in Section~\ref{software-architecture}.

Pose perturbations model uncertainty in the initial palm placement and its effect on the contacts formed during closure. For command \(a\), let \((R_a,t_a)\) denote the standoff pose: the initial palm pose in the object frame before motion toward the object. Pose perturbations comprise independent zero-mean isotropic Gaussian increments \(\delta p\sim\mathcal N(0,\sigma_p^2I_3)\) and \(\delta\phi\sim\mathcal N(0,\sigma_\phi^2I_3)\), where \(\sigma_p\) is in millimeters and the configured angular scale is converted from degrees to radians. With \(\xi=(\delta p,\delta\phi)\), the perturbed standoff is \((\exp([\delta\phi]_\times)R_a,t_a+\delta p)\), where \([\cdot]_\times\) maps a vector to its cross-product matrix. The travel direction undergoes the same incremental rotation. Approach and closure are rerun to obtain \(\mathcal C(a,\xi)\), then quality is evaluated at each equally weighted friction node. Thus pose uncertainty changes the contact set, and friction uncertainty changes the wrench hull at each sampled closure. The retained array includes the nominal pose and the configured number of perturbed approaches.

\subsubsection{Palm Frame and Discrete Approach Search}
\label{sec:hand-frame-construction}

The \texttt{HandFrame} specifies how an approach direction and roll determine a palm placement. Its nonzero approach axis \(a_H\), finger axis \(f_H\), and grasp-center point \(g_H\) are expressed in the palm frame, with \(g_H\) in millimeters. For nonparallel axes, the construction removes the component of \(f_H\) along \(a_H\) and forms the right-handed orthonormal basis
\begin{equation}
 \begin{aligned}
  f_H^{\perp}&=f_H-a_H\frac{f_H^\top a_H}{a_H^\top a_H},
  &\hat a_H&=\frac{a_H}{\|a_H\|_2},
  &\hat f_H&=\frac{f_H^{\perp}}{\|f_H^{\perp}\|_2},\\
  B_H&=\bigl[\hat a_H\ \ \hat f_H\ \ \hat a_H\times\hat f_H\bigr].
 \end{aligned}
 \label{eq:hand-alignment-basis}
\end{equation}
The finger axis therefore fixes the transverse palm orientation after orthogonalization. The grasp-center point fixes the part of the hand placed at the prescribed standoff from the object center.

Let \(d\) be the unit approach direction in the object frame and \(\theta\) the roll angle in radians. The constructor chooses \(r=e_z\) for \(|d_z|<0.9\), and \(r=e_x\) otherwise, then sets
\begin{equation}
 \begin{aligned}
 v_0&=\frac{d\times r}{\|d\times r\|_2},
 &v_\theta&=\operatorname{Rot}(d,\theta)v_0,
 &B_O&=[d\ \ v_\theta\ \ d\times v_\theta],\\
 R&=B_OB_H^\top,
 &t&=c-sd-Rg_H,
 &{}^{O}T_H&=\begin{bmatrix}R&t\\0&1\end{bmatrix},
 \end{aligned}
 \label{eq:standoff-construction}
\end{equation}
where \(s>0\) is the standoff in millimeters and \(\operatorname{Rot}(d,\theta)\) rotates about \(d\). Consequently, \(R\hat a_H=d\), \(R\hat f_H=v_\theta\), and \(Rg_H+t=c-sd\). Changing roll changes palm orientation while preserving the grasp-center location. The public \texttt{palm\_pose\_for\_approach} constructor takes roll in degrees and supplies the same transform to an external arm reachability calculation. Native placement composes it with the object pose, \({}^{W}T_H={}^{W}T_O{}^{O}T_H\), using the object pose captured when the planner is constructed.

The finite command bank combines \(J\) approach/roll pairs, \(P\) preshapes, and \(C\) closing subsets, giving \(N_{\mathrm{cmd}}=JPC\) grasp commands. An explicit approach already includes its roll. The generated bank instead combines each direction with equally spaced rolls \(2\pi j/n_{\mathrm{roll}}\). For \(n_{\mathrm{dir}}\) directions, its Fibonacci construction uses
\begin{equation}
 \begin{aligned}
 i_k&=k+\tfrac12,&
 \phi_k&=\arccos(1-2i_k/n_{\mathrm{dir}}),&
 \vartheta_k&=\pi(1+\sqrt5)i_k,\\
 d_k&=(\cos\vartheta_k\sin\phi_k,
       \sin\vartheta_k\sin\phi_k,\cos\phi_k),
 &k&=0,\ldots,n_{\mathrm{dir}}-1.
 \end{aligned}
 \label{eq:generated-approaches}
\end{equation}
Thus \(J=n_{\mathrm{dir}}n_{\mathrm{roll}}\) for the generated bank. Preshapes assign initial independent coordinates; closing subsets select which coordinates move during closure. With no supplied preshapes the planner uses the configured open posture, and the default closing choice enables all independent coordinates.

For each grasp command, approach resets the preshape and checks its standoff placement for collision. At increment \(h>0\), subsequent collision queries test translations \(t+jhd\). If the first scene collision occurs at step \(j\), the accepted translation is
\begin{equation}
 t_{\mathrm{pre}}=t+\max\{0,(j-1)h-\delta\}\,d,
 \label{eq:discrete-approach-clearance}
\end{equation}
where \(\delta\ge0\) is the clearance in millimeters. A colliding start or exhaustion of the travel limit without collision rejects the approach. The first detected collision can involve any scene geometry. Closure resets the preshape, places the palm at the accepted transform, checks collision, and runs autograsp closure with unselected coordinate velocities set to zero. After restoring the velocities, it refreshes the grasp, extracts object-side contacts, and applies the contact-count and contact-region tests. Admitted nominal closures must also pass the support-plane and keep-out tests on their final hand vertices.

\subsubsection{Planner Algorithm and Computational Work}
\label{sec:planner-work-counts}

The search separates acquisition of contact geometry from evaluation of its friction-dependent wrench hulls. The following algorithm preserves the accepted nominal placement, closed independent coordinates, and contacts while later native operations acquire perturbed closures.
\begin{enumerate}
 \item Enumerate each grasp command and construct its standoff transform. Run discrete approach, closure, and nominal feasibility tests; skip the command on failure.
 \item Save the nominal closure and initialize its random stream from the configured seed and command descriptor.
 \item For each pose perturbation, perturb the original standoff and travel direction, then repeat approach and closure. Apply contact-count and contact-region tests to the new contacts. Assign a zero-quality row to a failed scenario.
 \item At every friction node, score the nominal contacts and each successful perturbed contact set. Retain the complete \(\varepsilon\) array and, in task mode, the task-capacity array.
 \item Aggregate the configured score over all pose/friction cells using \eqref{eq:weighted-tail}; create the grasp record with the saved nominal closure and sampled scores.
 \item Sort all admitted records by decreasing score and return the requested prefix, or the complete list when no return limit is specified.
\end{enumerate}

For \(n_\xi\) perturbed approaches and \(n_\mu>0\) friction nodes, the array represents the empirical product law
\begin{equation}
 \widehat{\mathbb P}=
 \frac{\delta_0+\sum_{r=1}^{n_\xi}\delta_{\xi_r}}{n_\xi+1}
 \ \otimes\
 \frac{1}{n_\mu}\sum_{j=1}^{n_\mu}\delta_{\mu_j},
 \label{eq:empirical-pose-friction-law}
\end{equation}
where \(\delta_x\) denotes a unit point mass at \(x\), and \(\xi=0\) is the nominal pose. The nominal row has the same mass as each perturbed row. A failed perturbed closure retains its mass at zero quality across every friction column; it does not remove a scenario from the distribution. The stored nominal closure remains the reconstruction target even as perturbation evaluation changes the native scene. The caller applies the retained palm pose and closed independent coordinates to display or reconstruct the selected grasp.

Let \(N_{\mathrm{feas}}\) denote the number of commands passing all nominal filters. Every enumerated command invokes nominal approach, and every admitted command invokes another \(n_\xi\) approaches, so
\begin{equation}
 \begin{aligned}
 N_{\mathrm{approach}}&=N_{\mathrm{cmd}}+N_{\mathrm{feas}}n_\xi,\\
 N_{\mathrm{autograsp}}&\le N_{\mathrm{close}}
 \le N_{\mathrm{approach}}.
 \end{aligned}
 \label{eq:planner-native-work}
\end{equation}
In exact arithmetic, an approach makes at most \(1+\lceil2s/h\rceil\) scene collision queries, including the initial placement; a subsequent close makes its own placement query. Floating-point travel accumulation can alter a boundary step. Friction nodes add numerical scoring without repeating native closure. If \(n_{\mathrm{ok}}(a)\) counts successful perturbed closures for admitted command \(a\), the number of \texttt{analyze\_gws} calls is
\begin{equation}
 n_\mu\sum_{a\in\mathcal A}\bigl(1+n_{\mathrm{ok}}(a)\bigr)
 \le N_{\mathrm{feas}}(n_\xi+1)n_\mu.
 \label{eq:planner-numerical-work}
\end{equation}
Each call computes recomputed quality \(\varepsilon\) and volume; task mode additionally computes task-capacity hulls. The expression therefore counts analysis calls, independently of the number of internal hull calculations.

The retained \(\varepsilon\) arrays contain \(N_{\mathrm{feas}}(n_\xi+1)n_\mu\) scalars, and task mode retains the same number of additional capacity values. Their storage is \(O(N_{\mathrm{feas}}(n_\xi+1)n_\mu)\), in addition to nominal geometry and record data. Sorting requires \(O(N_{\mathrm{feas}}\log N_{\mathrm{feas}})\) comparisons. The \texttt{top\_k} argument limits the returned records after every admitted command has been scored. These counts distinguish increasing the geometric search bank, increasing pose reacquisition, and refining friction analysis as separate computational choices.

\subsubsection{Contact Wrench Models and Capacity}\label{sec:contact-wrench-models}

The wrench analysis links the contacts formed by closure to the external loads supported by the grasp. We model each contact \(i\) as a frictional point contact with position \(p_i\) and inward unit normal \(n_i\) in the object frame \citep{murray1994mathematical,rimon2019mechanics}. For contact force \(f\) and friction coefficient \(\mu\ge0\), the Coulomb conditions are \(f_n=n_i^\top f\ge0\) and \(\|f-f_n n_i\|_2\le\mu f_n\). The default cone approximation uses eight equally spaced unit tangent directions \(s_{ij}\) and edge forces \(f_{ij}=n_i+\mu s_{ij}\). Their moments about the center of mass \(c\) give the normalized wrench
\begin{equation}
 w_{ij}=\bigl(f_{ij},\,(p_i-c)\times f_{ij}/\rho\bigr).
 \label{eq:contact-wrench}
\end{equation}
Since each edge force has unit normal component, the sum of the nonnegative cone-edge coefficients is the total normal force. The wrench hull \(W=\operatorname{conv}(\{w_{ij}\})\) contains the wrenches obtainable with unit total normal force \citep{ferrari1992planning,miller_graspit_2004}. Force closure requires the origin in the interior of \(W\); under this condition, the hull represents a total normal-force cap of one. Recomputed Ferrari--Canny quality \(\varepsilon\) is the radius of the largest origin-centered Euclidean ball in \(W\). The positive length \(\rho\) scales torque relative to force. For recomputed quality \(\varepsilon\), the length is \(\rho_\varepsilon=\max\{\max_i\|p_i-c\|_2,0.01\,\mathrm{mm}\}\). For task-wrench capacity \(Q\), the hull and required loads share \(\rho_Q=100\)~mm. Thus \(\rho\) denotes the chosen metric-specific torque scale, with \(\rho_\varepsilon\) and \(\rho_Q\) identifying the two implemented conventions.

For each of \(K\) specified external task forces, we normalize \(f_k\) to unit length and retain its application point \(x_k\) in object-frame millimeters. The required balancing contact wrench is \(\ell_k=-\bigl(f_k,(x_k-c)\times f_k/\rho\bigr)\), about the same center of mass. Accordingly, the caller supplies the negated external-force direction to the planner's balancing-force input. The capacity is the supported external-force magnitude per unit total normal force along the least-supported required direction \citep{li1988task,borst2004grasp},
\begin{equation}
 Q(a,\xi,\mu) = \min_k \sup\{\lambda\geq 0:\lambda\ell_k\in W\}.
 \label{eq:task-quality}
\end{equation}
A failed closure or a realized grasp without force closure receives \(Q=0\).

The numerical interface also exposes the min-weight force-closure formulation used in FRoGGeR \citep{li2023frogger}. For \(W_m=[w_1\ \cdots\ w_N]\), the linear program is
\begin{equation}
 \ell^*=\max_{\alpha\in\mathbb R^N,\,\ell\in\mathbb R}\ \ell
 \quad\text{subject to}\quad
 W_m\alpha=0,\quad \boldsymbol{1}^{\top}\alpha=1,\quad
 \alpha_j\ge\ell\quad(j=1,\ldots,N).
 \label{eq:min-weight}
\end{equation}
A positive optimum supplies a strictly positive convex representation of the origin. Together, \(\ell^*>0\) and \(\operatorname{rank}W_m=6\) are equivalent to \(0\in\operatorname{int}W\), the geometric force-closure condition. The rank qualification excludes balance confined to a lower-dimensional wrench subspace. The SciPy implementation returns the optimum and a Boolean requiring both an optimum above the numerical tolerance and wrench-matrix rank six. Retaining the optimum independently of the rank test distinguishes a positive margin within a lower-dimensional wrench subspace from six-dimensional force closure. The optional SciPy-free calculation returns the minimum-singular-value proxy. The min-weight coefficient, Ferrari--Canny radius, and task-wrench capacity therefore retain distinct definitions and numerical outputs.

For task-wrench capacity, the convex-hull membership condition in \eqref{eq:task-quality} also has a coefficient-allocation formulation. Using the wrench matrix \(W_m\) defined in \eqref{eq:min-weight}, the capacity along a required balancing wrench \(\ell_k\) is
\begin{equation}
 q_k^{\mathrm{ray}}=
 \max_{\alpha\in\mathbb R^N_{\ge0},\,\lambda\ge0}\lambda
 \quad\text{subject to}\quad
 W_m\alpha=\lambda\ell_k,\qquad
 \boldsymbol{1}^{\top}\alpha=1.
 \label{eq:ray-allocation}
\end{equation}
The equality on the coefficients imposes one total normal-force budget across all contacts. With strict six-dimensional force closure, \(Q=\min_k q_k^{\mathrm{ray}}\). This formulation computes the same directional capacity through linear programs without constructing the hull facets. We use the rank-qualified min-weight certificate to establish strict closure before evaluating the directional programs. The numerical implementation equilibrates the wrench-coordinate equations by applying the same invertible row scaling to the wrench matrix and task loads. Task magnitudes are normalized during each solve and restored in the returned multiplier. Scaled equality and inequality residuals are checked against a tolerance of \(10^{-9}\). For Ferrari--Canny quality, the supporting-plane calculation retains its Euclidean definition. A deterministic hull retry handles degenerate facet topology without perturbing the input wrenches, with checks of finite normalized planes, containment of the input points, and strict exclusion of an origin on a supporting plane.

\subsubsection{Force Budgets and Wrench-Metric Comparisons}
\label{sec:wrench-normalization}

Under the contact model, load capacity depends on the permitted contact-force budget. Primitive-wrench normalization determines the budget represented by the wrench hull. The array builder \texttt{build\_grasp\_}\allowbreak\texttt{matrix} exposes this choice through \texttt{unit\_normal\_force}, whose default is \texttt{True}. Under the unit-normal-force convention in \eqref{eq:contact-wrench}, nonnegative edge coefficients \(\alpha_{ij}\) produce contact forces \(f_i=\sum_j\alpha_{ij}f_{ij}\). For unit normals orthogonal to their unit tangent directions,
\begin{equation}
 \sum_i n_i^\top f_i=\sum_{i,j}\alpha_{ij}.
 \label{eq:normal-force-budget}
\end{equation}
The convex-hull coefficient constraint therefore fixes total normal force at unity across all contacts. If \(0\in W\), replacing the coefficient equality by a cap gives \(\operatorname{conv}(\{0\}\cup W)=W\). Strict force closure supplies this origin-membership condition. The budget is shared across contacts, so it does not impose separate normal-force limits at individual contacts.

Setting \texttt{unit\_normal\_force=False} divides each primitive force and moment by \(\sqrt{1+\mu^2}\). Let superscripts \(\mathrm n\) and \(\mathrm e\) denote the unit-normal and unit-edge conventions, respectively, and define \(k(\mu)=1/\sqrt{1+\mu^2}\). At one common scalar friction coefficient,
\begin{equation}
 W_m^{\mathrm e}=k(\mu)W_m^{\mathrm n},\qquad
 W^{\mathrm e}=k(\mu)W^{\mathrm n}.
 \label{eq:edge-normalization-dilation}
\end{equation}
In ideal orthonormal contact frames, every alternative primitive force has unit Euclidean norm and normal component \(k(\mu)\). A unit sum of nonnegative coefficients consequently fixes total normal force at \(k(\mu)\), while \(\sum_i\|f_i\|_2\le1\) by the triangle inequality. Equality of resultant contact-force norms need not hold because tangential components of several edges at one contact can cancel. The alternative hull thus represents a primitive-edge coefficient budget bounding the sum of resultant contact-force norms.

A common positive dilation preserves origin interior and matrix rank, so both conventions have the same geometric force-closure status at fixed \(\mu\). The min-weight linear program in \eqref{eq:min-weight} has a stronger coordinate invariance: for any invertible wrench-coordinate matrix \(D\), the equations \(DW_m\alpha=0\) and \(W_m\alpha=0\) have identical solutions. Thus
\begin{equation}
 \ell^*(DW_m)=\ell^*(W_m),\qquad
 \ell^*(W_m)\le\frac{1}{N},
 \label{eq:min-weight-invariance-bound}
\end{equation}
where the bound holds for a feasible program and follows by summing \(\alpha_j\ge\ell\) over the \(N\) columns. The optimum measures a coefficient margin and depends on the generator representation: duplicated or redundant columns can change it without changing the hull. The invariance applies to the linear program; a singular-value proxy scales under a uniform dilation. Absolute numerical thresholds can also affect computed closure predicates near degeneracy.

Geometric scores respond differently to a change in force budget. For a positive factor \(k\), fixed task loads, and strict origin interior,
\begin{equation}
 \varepsilon(kW)=k\varepsilon(W),\qquad
 Q(kW,\{\ell_k\})=kQ(W,\{\ell_k\}),\qquad
 \operatorname{vol}_6(kW)=k^6\operatorname{vol}_6(W).
 \label{eq:wrench-score-dilation}
\end{equation}
Here the explicit hull and load arguments of \(Q\) indicate the membership problem in \eqref{eq:task-quality}. The capacity law follows from \(\lambda\ell_k\in kW\) exactly when \((\lambda/k)\ell_k\in W\). Applying the same invertible coordinate transformation to both the hull and required loads instead preserves capacity. The common torque rescaling discussed below is such a transformation. A general anisotropic transformation changes Euclidean balls, so it need not preserve Ferrari--Canny quality. Changing only the primitive normalization changes the force budget against unchanged loads.

With unit-normal-force generators, increasing friction enlarges the attainable wrench hull for fixed contacts. Under the alternative convention, capacity becomes \(k(\mu)Q^{\mathrm n}(\mu)\); an increasing unit-normal capacity multiplied by decreasing \(k(\mu)\) has no general monotonicity implication. At a single common friction coefficient, the factor \(k(\mu)\) scales Ferrari--Canny qualities and task capacities of all fixed geometries equally, preserving their order. Across friction nodes, the factor varies by column, so mean and lower-tail rankings need not be preserved. The linear-program min-weight optimum remains invariant separately at each node. The planner uses the default unit-normal convention; comparison with the alternative convention is available through caller-defined array analysis.

For fixed contacts, increasing the friction coefficient enlarges each polygon of admissible unit-normal forces. The convex hull of the equally spaced tangent directions \(s_{ij}\) contains the origin, so scaling the polygon by a larger nonnegative factor preserves every point in the smaller polygon. The linear force-to-wrench map and convex hull over contacts preserve this inclusion. Hence \(\mu_1\le\mu_2\) implies inclusion of the corresponding wrench hulls and \(Q(a,\xi,\mu_1)\le Q(a,\xi,\mu_2)\) for the fixed closure. A common invertible torque rescaling \(D=\operatorname{diag}(I_3,\gamma I_3)\), \(\gamma>0\), satisfies \(\lambda\ell_k\in W\) exactly when \(\lambda D\ell_k\in DW\). Task-wrench capacity is consequently invariant to the common torque scale, while the Ferrari--Canny ball depends on this scale through its Euclidean norm.

\subsubsection{Lower-Tail Aggregation of Quality}\label{sec:lower-tail-aggregation}

Let \(Z_a=Q(a,\xi,\mu)\) over the selected pose and friction distribution; recomputed \(\varepsilon\) can be used in place of \(Q\). The planner selects a feasible command by maximizing lower-tail conditional value-at-risk for mass \(\beta\in(0,1]\) \citep{rockafellar2000cvar},
\begin{equation}
 a^\star\in\arg\max_{a\in\mathcal A} \sup_{\eta\in\mathbb R}
  \left\{\eta-\frac{1}{\beta}\mathbb E[(\eta-Z_a)_+]\right\},
 \label{eq:tail-selection}
\end{equation}
where \(\mathcal A\) contains the commands passing the nominal feasibility filters and \((x)_+=\max(x,0)\). The objective is the mean at \(\beta=1\), while nominal scoring evaluates \(Q(a,0,\mu_0)\) at reference friction \(\mu_0\). For weighted samples \(z_1,\ldots,z_M\) with normalized probabilities \(\omega_i\ge0\), let \(z_{(1)}\le\cdots\le z_{(M)}\), and let \(\omega_{(i)}\) denote the corresponding weights. With \(\Omega_i=\sum_{j=1}^i\omega_{(j)}\) and \(\Omega_0=0\), the empirical objective is
\begin{equation}
 \widehat{\operatorname{CVaR}}_\beta
 =\frac{1}{\beta}\sum_{i=1}^M
 z_{(i)}\min\{\omega_{(i)},(\beta-\Omega_{i-1})_+\}.
 \label{eq:weighted-tail}
\end{equation}
The sorted samples contribute from lowest quality upward until their cumulative probability reaches \(\beta\). If the last included sample would exceed this mass, only the required fraction of its weight enters the average. Equal weights yield the lowest \(\beta M\) samples with a fractional final sample. When the zero-quality scenarios defined in \eqref{eq:empirical-pose-friction-law} fill the lower tail, every affected command receives an identical zero score and their tail-based ranking is tied. Large costs and small grasp qualities are adverse outcomes. Negating quality maps its lower tail to the upper tail of a cost, so the numerical interface computes lower-tail quality CVaR as \(-\operatorname{CVaR}^{\mathrm{cost}}_\beta(-z)\).

Averaging quality over an adverse friction tail is distinct from evaluating quality at an average tail friction. For fixed contacts and nondecreasing quality \(g(\mu)\),
\begin{equation}
 \operatorname{CVaR}_\beta[g(\mu)]
 =\frac{1}{\beta}\int_0^\beta g\!\left(F_\mu^{-1}(u)\right)\,du,
 \label{eq:friction-tail}
\end{equation}
where \(F_\mu^{-1}\) is the friction quantile function. Evaluating the quality at tail-average friction gives \(g(\operatorname{CVaR}_\beta[\mu])\); equality with \eqref{eq:friction-tail} holds for affine \(g\) on the tail interval. The sampled objective evaluates quality before averaging and also includes the contact changes produced by pose perturbations. Additional objectives can be applied to the saved arrays without repeating their closures.

\subsection{Export, Closure, and Grasp Reconstruction}\label{sec:runnable-workflow}
The executable workflow exports the RealHand L6 and a 50~mm cube from a compiled MuJoCo model, searches for grasp commands in GraspIt!, and reconstructs the selected command in MuJoCo. Listing~\ref{lst:workflow} builds the wrapper image and runs the public example. The distributed Dockerfile uses Ubuntu~24.04 with Qt~5, Coin3D/SoQt, Qhull, pybind11, NumPy, SciPy, and MuJoCo. Its default preparation pins an upstream GraspIt! source snapshot at commit \texttt{de64377} and applies the supplied compatibility patch. The \href{https://github.com/coenwerem/graspit-py/blob/0552782d489b2d1542edfa4b65b3d6a9f9a6802c/docker/prepare_graspit.sh}{engine-preparation script} records the complete commit identifier. The interface measurements in Section~\ref{sec:ros-benchmark} identify their evaluated engine revision separately. The variables \texttt{WRAPPER} and \texttt{HAND\_ASSETS} identify the wrapper checkout and the external RealHand L6 mesh directory. We configured the search with 40 approach directions, six rolls, and six preshapes. Recomputed quality \(\varepsilon\) is evaluated for the nominal pose and four perturbed approaches at friction nodes 0.4, 0.7, and 1.0, giving fifteen equally weighted pose--friction values per command. The output contains robot and object XML, link meshes, the joint map, and up to five ranked grasp records. The evaluated instance returned five records; its highest-ranked grasp had eight contacts and empirical lower-tail CVaR \(\widehat{\operatorname{CVaR}}_{0.5}(\varepsilon)=0.00896\) for recomputed quality \(\varepsilon\).

\noindent\begin{minipage}{\linewidth}
\begin{lstlisting}[language=bash,caption={Plan a RealHand L6 Grasp and Reconstruct Its MuJoCo Configuration},label={lst:workflow}]
docker build -f "$WRAPPER/docker/Dockerfile" -t graspit-py "$WRAPPER"
bash "$WRAPPER/examples/l6_workflow/run.sh" graspit-py \
  "$HAND_ASSETS" ./l6-workflow-output
\end{lstlisting}
\end{minipage}

Listing~\ref{lst:record-reconstruction} shows how the public workflow reads a grasp record and restores source hinge coordinates. The initialized planner uses the exported hand and object with the supplied search configuration. The manifest maps independent coordinates to named source hinges, while the record retains object-frame contact positions and inward normals for numerical analysis.

\noindent\begin{minipage}{\linewidth}
\begin{lstlisting}[language=Python,caption={Read a Grasp Record and Restore the Source Hinge Coordinates},label={lst:record-reconstruction}]
import mujoco as mj
import numpy as np

selected = planner.plan(top_k=5)[0].to_dict()
T_OH = np.asarray(selected["T_object_hand_mm"])
contacts_O = selected["contacts"]
for chain in manifest["chains"]:
    for joint in chain:
        u = selected["dofs_rad"][joint["dof"]]
        data.joint(joint["mujoco_joint"]).qpos[0] = (
            joint["ratio"] * u + joint["offset_rad"])
mj.mj_forward(model, data)
\end{lstlisting}
\end{minipage}

Each source hinge angle is reconstructed from the independent coordinates using \eqref{eq:joint-coupling} and assigned by its MuJoCo joint name. With the palm fixed by the robot configuration, the workflow places the object so the object-relative palm pose equals the planned transform, converting translation from millimeters to meters. Forward kinematics then recovers the requested relative placement. The translation and rotation residuals were at floating-point precision. This is a kinematic reconstruction of the command, which an external simulator can subsequently use for dynamic execution.

\section{Usage and Research Workflows}\label{sec:research-workflows}

The contact and coordinate representations support three research tasks: evaluating external grasp proposals, constraining contacts for tool use or assembly, and analyzing saved scores. These workflows connect the mathematical definitions to caller-supplied models, commands, and uncertainty distributions through executable interface operations.

\subsection{Evaluating Externally Supplied Grasp Proposals}\label{sec:proposal-workflow}

An external grasp generator supplies a palm pose, preshape, and desired closing posture, while the wrapper supplies a common native model for closure and contact extraction. A learned proposal generator or contact optimizer can therefore use the native interface without adopting the supplied planner's command bank. The caller first converts the proposed palm placement to the world-frame convention and checks the assigned preshape for scene collisions. Collision-aware movement then advances toward the caller's closing coordinates. Listing~\ref{lst:proposal-contacts} retains the coordinate trace and refreshes the selected object's grasp before acquiring common-frame contacts. All poses, coordinate indices, and closing targets come from the caller's loaded hand and scene.

\noindent\begin{minipage}{\linewidth}
\begin{lstlisting}[language=Python,caption={Acquire Contacts from a Caller-Supplied Proposal},label={lst:proposal-contacts}]
from graspit_py.core.bindings import Grasp

world.set_current_hand_pose(
    palm_quat_wxyz, palm_translation_mm)
world.set_current_hand_dof_values(preshape_rad)
if not world.no_collision():
    raise ValueError("Preshape intersects scene geometry")
moved, trace = world.trace_move_current_hand_dof_to_contacts(
    closing_target_rad, desired_step_sizes=closing_steps_rad,
    stop_at_contact=closure_stop_at_contact)
grasp = Grasp(world, graspable_body_index=object_index)
if not grasp.compute():
    raise ValueError("Closure has no valid native grasp")
contacts_O = grasp.object_contact_geometry()
c_O_mm = world.get_graspable_body_cog_local(object_index)
\end{lstlisting}
\end{minipage}

Returning contact positions and normals allows the numerical objective to vary independently of the proposal generator. Listing~\ref{lst:contact-analysis} builds \(W_m\) across caller-selected friction coefficients using object-frame contacts and center of mass, with an explicit common torque scale \(\rho\). The returned min-weight optimum and force-closure predicate retain separate meanings. This sweep holds the realized contact geometry fixed; the friction-sampling helper similarly analyzes fixed contacts, while the planner's pose samples rerun approach and closure. Caller-supplied query points and object triangles also permit unsigned geometric-distance analysis in the same frame and length unit. A new contact model, load direction, or friction grid requires new numerical scores; changing only the aggregation of saved scores does not require another closure.

\noindent\begin{minipage}{\linewidth}
\begin{lstlisting}[language=Python,caption={Analyze Fixed Contacts and Object Geometry},label={lst:contact-analysis}]
import numpy as np
from graspit_py.grasp.quality import (
    build_grasp_matrix, min_weight_force_closure)
from graspit_py.utils.geometry import compute_distance_to_mesh

positions_O_mm = np.asarray(
    [contact["position_mm"] for contact in contacts_O])
normals_O = np.asarray([contact["normal"] for contact in contacts_O])
friction_results = []
for mu in friction_values:
    W_m = build_grasp_matrix(
        positions_O_mm, normals_O, c_O_mm,
        friction_coef=mu, max_radius=rho_mm)
    l_star, force_closure = min_weight_force_closure(W_m)
    friction_results.append((mu, l_star, force_closure))
distances_mm = compute_distance_to_mesh(
    query_points_O_mm, object_vertices_O_mm, object_faces)
\end{lstlisting}
\end{minipage}

\subsection{Sampling Friction for Fixed Contacts}
\label{sec:fixed-contact-friction}

The function \texttt{compute\_gws\_under\_friction\_uncertainty} evaluates friction sensitivity of an acquired contact set. Contact positions, inward normals, and object center remain fixed while each friction draw produces a new wrench analysis. The helper samples \(X\sim\mathcal N(m_\mu,\sigma_\mu^2)\) and uses \(\mu=\operatorname{clip}(X,\mu_-,\mu_+)\), with \(\mu_-=0.1\) and \(\mu_+=2.0\). Its \texttt{friction\_mean} and \texttt{friction\_std} arguments specify \(m_\mu\) and \(\sigma_\mu\), respectively. Each draw calls \texttt{analyze\_gws} with default unit-normal-force generators and the torque scale \(\rho_\varepsilon\) defined from the maximum contact distance in Section~\ref{sec:contact-wrench-models}. The computation requires no native approach or closure, so its sampled variation is conditional on the supplied contact geometry.

Clipping produces endpoint probability masses as well as a continuous interior density. For \(\sigma_\mu>0\), with \(\Phi\) denoting the standard normal cumulative distribution function,
\begin{equation}
 \mathbb P(\mu=\mu_-)=\Phi\!\left(\frac{\mu_--m_\mu}{\sigma_\mu}\right),\qquad
 \mathbb P(\mu=\mu_+)=1-\Phi\!\left(\frac{\mu_+-m_\mu}{\sigma_\mu}\right).
 \label{eq:clipped-friction-atoms}
\end{equation}
Inside the interval the density equals the original Gaussian density. Consequently, the supplied Gaussian parameters generally differ from the mean and standard deviation of the clipped distribution. At \(\sigma_\mu=0\), all mass is concentrated at \(\operatorname{clip}(m_\mu,\mu_-,\mu_+)\).

The returned \texttt{epsilon\_samples} and \texttt{volume\_samples} arrays contain one value per draw. The dictionary also provides their means and population standard deviations, the minimum and maximum recomputed quality, the sample count, and \texttt{fc\_probability}. When SciPy successfully constructs a six-dimensional hull, the volume values are \(10^6\operatorname{vol}_6(W)\), a scaled six-dimensional normalized-wrench-hull volume. The helper returns zero for handled hull-construction failures. Their standard deviation describes sampled variation; it is not a standard error. The returned \texttt{fc\_probability} field is the fraction of friction draws classified as force closure by \texttt{analyze\_gws} under the installed numerical backend. It describes friction-dependent numerical classification of the fixed contacts, not a frequency of successful grasp acquisition. In particular, the helper's predicate is separate from the rank-qualified min-weight test.

Listing~\ref{lst:fixed-contact-friction} supplies a seeded NumPy generator and computes a lower-tail statistic from the returned quality array. The helper consumes the supplied generator state and returns score samples without the sampled friction values. Additional tail masses can be evaluated from the saved qualities without repeating analysis. A study requiring scores at specified friction values can use an explicit grid, as in Listing~\ref{lst:contact-analysis}. Planner arrays additionally include contact changes from pose perturbations and retain the nominal pose under \eqref{eq:empirical-pose-friction-law}.

\noindent\begin{minipage}{\linewidth}
\begin{lstlisting}[language=Python,caption={Sample Friction for an Acquired Contact Set},label={lst:fixed-contact-friction}]
import numpy as np
from graspit_py.core.types import ContactPoint
from graspit_py.grasp.quality import (
    compute_gws_under_friction_uncertainty)
from graspit_py.risk import lower_tail_mean

contacts = [ContactPoint(np.asarray(c["position_mm"]),
                         np.asarray(c["normal"]))
            for c in contacts_O]
summary = compute_gws_under_friction_uncertainty(
    contacts, c_O_mm,
    friction_mean=friction_mean,
    friction_std=friction_std,
    n_samples=n_friction_samples,
    rng=np.random.default_rng(seed))
epsilon_samples = summary["epsilon_samples"]
tail_quality = lower_tail_mean(epsilon_samples, alpha=beta)
\end{lstlisting}
\end{minipage}

\subsection{Configuring Tool and Fixture Constraints}\label{sec:fixture-workflow}

A tool-use or assembly study specifies an allowed contact region on the object while preserving access to its working end or surrounding fixture. The caller expresses this region in object-frame millimeters and supplies preshapes and closing subsets appropriate to the hand. Holding selected independent coordinates at their preshape values permits the remaining coordinates to close around the allowed region. Contact-region tests constrain the acquired contacts; nominal support and keep-out predicates constrain the final hand vertices. A keep-out box excludes points strictly inside it; a hand vertex exactly on its boundary is accepted. Because the box test uses vertices, edge and face intersections require the caller's additional geometric checks. The standoff transform constructor supplies the same approach placement to an external arm reachability routine before command enumeration.

Task rows connect these geometric choices to a required load. Each row specifies an object-frame application point and a balancing-force direction, so the caller negates the direction of the external force to be resisted. The planner normalizes each direction individually; entered force magnitudes do not prescribe unequal load magnitudes. The resulting task-wrench capacity ranks resistance along the specified directions using the convention in Section~\ref{contact-based-planning}. Pose perturbations rerun approach and closure, allowing the score array to reflect contact changes and closure failures across friction nodes. Contact-count and contact-region tests apply to these perturbed closures; support and fixture predicates apply to the nominal final vertices. External arm planning supplies feasible robot motion for the resulting command targets.

\subsection{Inspecting Saved Scores and Source Correspondence}\label{sec:record-workflow}

Saved pose-by-friction arrays support analysis of commands whose nominal quality differs from their performance under modeled perturbations. A researcher examines weak pose rows or friction columns, then compares the mean with lower-tail averages at selected tail masses. Listing~\ref{lst:score-postprocessing} uses the retained Ferrari--Canny quality samples directly and accepts scenario weights aligned with the saved array. The standalone risk function normalizes supplied nonnegative weights and includes only the probability mass needed from the boundary sample. The stored search configuration identifies the pose distribution and friction nodes; a reweighted result describes the same sampled outcomes under the supplied weights.

\noindent\begin{minipage}{\linewidth}
\begin{lstlisting}[language=Python,caption={Reaggregate Saved Pose--Friction Scores},label={lst:score-postprocessing}]
import numpy as np
from graspit_py.risk import lower_tail_mean

samples = np.asarray(record["epsilon_samples"])
mean_quality = samples.mean()
tail_quality = lower_tail_mean(samples.ravel(), alpha=beta)
weighted_tail_quality = lower_tail_mean(
    samples.ravel(), alpha=beta,
    weights=np.asarray(scenario_weights).ravel())
\end{lstlisting}
\end{minipage}

The same record supports source-model reconstruction through a different representation: its accepted palm placement and closed independent coordinates. The exporter manifest expands each independent coordinate into named source hinges through \eqref{eq:joint-coupling}. Composing the object world transform with the recorded object-relative palm transform, after converting millimeters to meters, supplies a palm target to an external arm planner. Section~\ref{sec:runnable-workflow} provides the executable reconstruction example. For model diagnosis, the manifest's joint and link identities connect source hinges to native link poses and pure exported forward kinematics. Comparing these poses after the composition in \eqref{eq:api-link-composition} localizes coordinate or frame-conversion discrepancies without conflating them with the saved quality aggregation.

\section{Evaluation and Discussion}\label{verification}

\subsection{Geometry in Native
GraspIt!}\label{geometry-in-native-graspit}

We evaluated agreement between the exported GraspIt! link geometry and the MuJoCo source geometry across hand configurations. We assigned each sampled independent coordinate vector in native GraspIt! and read back the applied values. We then computed driven angles directly from the compiled MuJoCo equalities. Both export modes were evaluated at all 75 configurations of the RealHand L6, LEAP, Ability, Shadow, and Inspire models. Source-mode errors were computed from source-mesh vertices, and hull-mode errors from the vertices of each compiled collision hull. Corresponding sets were expressed in their respective palm frames. For finite vertex sets \(A\) and \(B\), the symmetric Hausdorff distance is
\begin{equation}
 d_H(A,B)=\max\left\{
 \max_{a\in A}\min_{b\in B}\|a-b\|_2,
 \max_{b\in B}\min_{a\in A}\|b-a\|_2
 \right\}.
 \label{eq:vertex-error}
\end{equation}
Table~\ref{tab:geometry-correspondence} reports the maximum over links and configurations. Across the five hands and both modes, the largest discrepancy was approximately 1.3~\(\mu\)m, compared with the 100~\(\mu\)m geometry tolerance, and requested and applied independent coordinates agreed. The L6, Ability, and Inspire models have coupled hinges; LEAP has independent hinges. The Shadow palm-and-finger planning model uses four declared equal-angle distal couplings, with its wrists outside the exported subtree.

We evaluated the reference correction and geometric approximations on synthetic models. A two-hinge model had driver and driven references of \(0.35\) and \(-0.20\)~rad and coefficients \(c_0=0.16\)~rad and \(c_1=0.70\). At three driver angles, the exported \(-0.285\)~rad offset satisfied the source equality within \(10^{-12}\)~rad. A three-finger model with nearly parallel axes, affine couplings, and a rigid distal body had a maximum vertex error of 14.4~\(\mu\)m over three configurations and seven links, also within the geometry tolerance. Hull-mode comparisons use compiled hull connectivity, while source-mode comparisons use source triangles. An independent MuJoCo collision-distance probe checked the outer surface of a concave mesh's convex hull at three coupled-joint configurations. We also verified connectivity to confirm that exported faces did not join vertices from distinct collision components.

\begin{table}[t]
\centering
\caption{\scshape Exported Link Geometry by Mesh Representation. Errors are maximum symmetric Hausdorff distances between corresponding palm-frame vertex sets. All tabulated discrepancies were within the 100~\(\mu\)m tolerance.}
\label{tab:geometry-correspondence}
\small
\begin{tabular}{lrrrrr}
\toprule
\rowcolor{TableHeader}
Hand & Coordinates & Links & Configurations & Source (\(\mu\)m) & Hull (\(\mu\)m) \\
\midrule
RealHand L6 & 6 & 12 & 49 & \(<0.01\) & \(<0.01\) \\
LEAP & 16 & 17 & 4 & \(<0.01\) & \(<0.01\) \\
Ability & 6 & 11 & 12 & \(<0.01\) & \(<0.01\) \\
Shadow, palm & 18 & 23 & 5 & \(<0.01\) & \(<0.01\) \\
Inspire, right & 6 & 13 & 5 & 0.836 & 1.29 \\
Two-hinge model & 1 & 3 & 3 & \(<0.01\) & \textemdash \\
\bottomrule
\end{tabular}
\end{table}

Figure~\ref{fig:export-embodiments} illustrates three configurations from the five-model geometry evaluation. Joint values were assigned by source name, and the exported link poses and meshes were queried in GraspIt!. Within each source/export pair, a common view shows the coupled index chains of the L6 and Ability hands and the independent LEAP coordinates. Palm-frame axes and finger labels identify the corresponding components; the reported per-model error covers the evaluated configurations.

\input{figures/export-embodiments.tex}

\subsection{Reference Angles and Collision Components}\label{sec:exporter-ablation}

The geometry correspondence depends on the affine joint map and the collision components attached to each link. We isolated their effects by omitting the reference-angle contribution in \eqref{eq:reference-coupling} or replacing a link's convex components with one hull over their combined vertices. The comparison used the five hand models in Table~\ref{tab:geometry-correspondence}, a coupled-joint model with nonzero reference angles, and a U-shaped component union with a prescribed gap. The two controlled models exercised the specific conversion mechanisms. For each model, we evaluated three configurations at the 0.25, 0.50, and 0.75 fractions of its feasible independent-coordinate intervals. Source surfaces and compiled hulls were evaluated with the complete and reference-ablated exporters, and component merging was evaluated for compiled hulls. The resulting 105 configuration comparisons included 42 complete-export configurations. Independent coordinates, joint limits, source geometry, and compiler settings were held fixed across the variants.

Every complete-export configuration satisfied the 0.1~mm correspondence tolerance without adjustment of assigned coordinates at joint limits. In the nonzero-reference model, the affine correction was \(-0.445\)~rad. Omitting this term displaced its link vertices by as much as 13.4~mm at the prescribed configurations. The reference-angle contribution was zero in the five hand scenes, so omitting it left their geometry unchanged. The controlled comparison therefore establishes the role of the reference term for models with nonzero reference angles, while the hand measurements establish correspondence for the evaluated source models.

Component merging changes the collision set even when the original component vertices remain inside the merged hull. We measured the added occupied region using \(21^3\) uniformly spaced cell centers in each link's component-union bounding box. Each point was classified against both the original union and the merged hull. Multiplying the number of points occupied only by the merged hull by the cell volume gave the added-volume estimate in Table~\ref{tab:component-ablation}. Merging introduced occupied regions in the Shadow and Inspire models. The L6, LEAP, and Ability scenes each had one compiled component per exported link, so their collision sets were unchanged. The complete exporter preserved the separate components in every case.

\begin{table}[t]
\centering
\caption{\scshape Effect of Combining Collision Components. Each link used \(21^3\) cell-center probes in its component-union bounding box. The percentage is the largest fraction of probes added on any link. Added volumes are midpoint-grid estimates summed over links. Italic entries give the occupied volume added by merging.}
\label{tab:component-ablation}
\small
\setlength{\tabcolsep}{5pt}
\begin{tabular}{lcrr}
\toprule
\rowcolor{TableHeader}
Model & \shortstack{Links with added probes\\/ compound links} & \shortstack{Maximum added\\probes (\%)} & \shortstack{Added volume\\(cm\(^3\))} \\
\midrule
Shadow & 1/5 & 25.8\% & \salmoncell{68.2} \\
Inspire & 1/1 & 13.7\% & \salmoncell{87.3} \\
U-shaped model & 1/1 & 56.0\% & \salmoncell{60.5} \\
\bottomrule
\end{tabular}
\end{table}

For the U-shaped model, the exact added volume was 64~cm\(^3\), and the fixed grid estimated 60.5~cm\(^3\). The prescribed sphere occupied the gap without interpenetration in the original component union. Combining the components filled the gap, produced eleven object contacts, and changed the final joint coordinate by 0.25~rad under the identical closing command.

\subsection{Native Hand Models and Reusable Profiles}\label{sec:hand-profiles}
The original Barrett model is supported through the native GraspIt! robot interface \citep{miller_graspit_2004}. We loaded the Barrett model, with four independent coordinates and nine links, and applied autograsp closure to a 50~mm cube centered at (0, 0, 100)~mm in the palm frame. The resulting grasp had six contacts and native Ferrari--Canny quality \(\varepsilon_{\mathrm{GI}}=0.146\). This route uses the native model's joint transmission and provides closure, contact geometry, and wrench metrics through Python.

Preparation and export profiles are provided for the Shadow and Inspire models evaluated in Table~\ref{tab:geometry-correspondence}. A profile identifies the palm root, independent coordinates, preshape, and affine joint relations. URDF collision meshes are resolved during preparation, and mimic relations are converted to MuJoCo equalities with the reference correction in \eqref{eq:reference-coupling}. The Inspire profile preserves six mimic relations and their angular offsets; Shadow's four equal-angle distal planning couplings are declared explicitly. Intersecting every driven hinge limit with its independent-coordinate interval gives the feasible intervals used to evaluate five interior configurations per mode, each applied without coordinate adjustment. The \href{https://clintonenwerem.com/graspit-py/hands.html}{hand-model guide} provides the source-model links, profiles, commands, coordinate conventions, and asset licenses.

\begin{figure}[!t]
\centering
\input{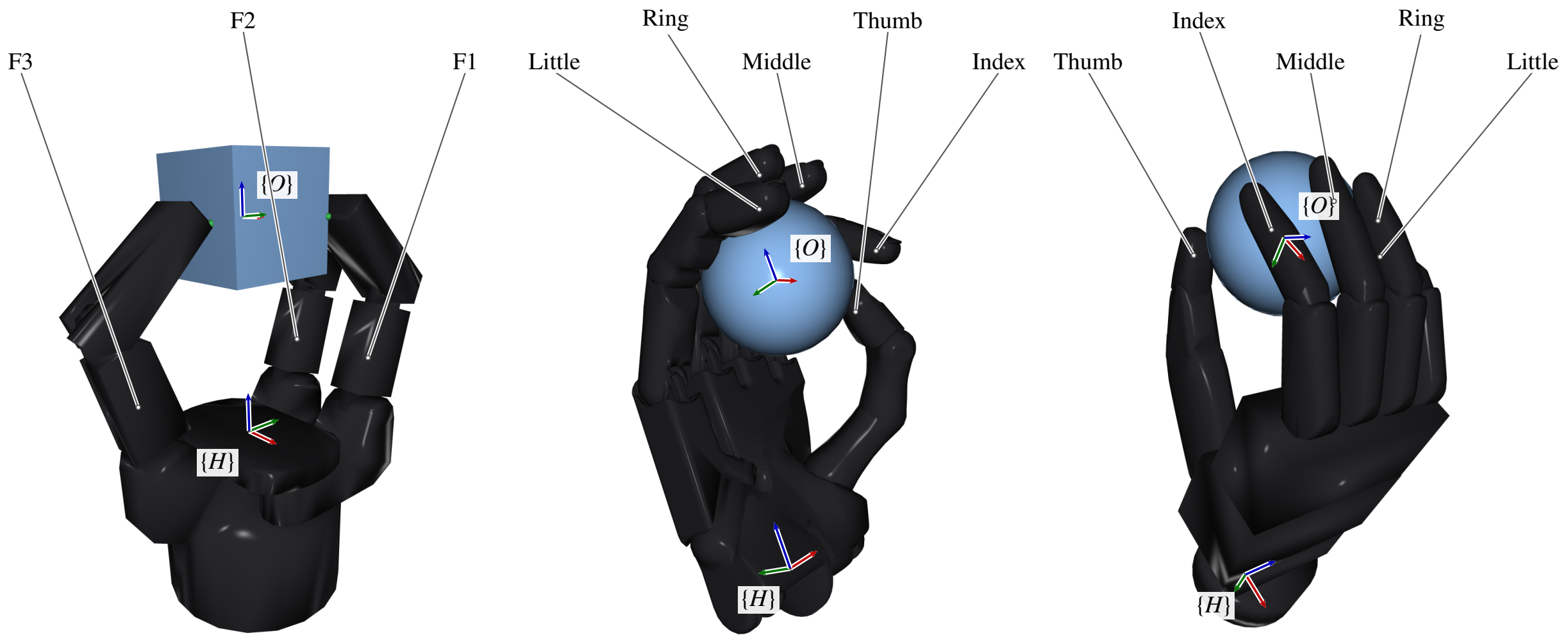}
\caption{\textbf{Grasps on Native and Exported Hand Models.} (a) Native Barrett closure on a 50~mm cube. (b, c) Exported Shadow and Inspire closures on a sphere of 60~mm diameter. Green markers indicate object-side contacts from native collision-based closure. Shadow is shown with its URDF visual meshes at the closed joint configuration. Its contact positions and quality were computed with the collision geometry. Palm frame \(\{H\}\), object frame \(\{O\}\), and finger labels are projected from the queried configuration. Red, green, and blue axes denote the respective frame's \(x\), \(y\), and \(z\) directions. F1--F3 identify Barrett's three kinematic finger chains. Per-panel readouts give the contact count and native Ferrari--Canny quality \(\varepsilon_{\mathrm{GI}}\).}
\label{fig:supported-hand-grasps}
\end{figure}

Closing the exported Shadow and Inspire models on a sphere of 60~mm diameter produced eleven and seven contacts, respectively, distributed over all five fingers. Their native qualities were \(\varepsilon_{\mathrm{GI}}=0.423\) and 0.539. None of the Barrett, Shadow, or Inspire closures involved palm contact. At the closed Shadow and Inspire configurations, the maximum compiled-model/export vertex discrepancy was below 0.01~\(\mu\)m. Figure~\ref{fig:supported-hand-grasps} shows the resulting grasp configurations, object-side contact positions, and palm and object frames.

\subsection{Independent Wrench-Mechanics Validation}\label{sec:numerical-validation}

The coefficient-allocation program in \eqref{eq:ray-allocation} depends on the cone-edge representation and the strict force-closure predicate. We evaluated its numerical agreement with a separate Cartesian-force formulation using the faces of each friction polygon. Let \(t_{i1},t_{i2}\) be orthonormal contact tangents and \(s\) the number of polygon edges. Define \(v_{ij}=\cos(2\pi(j+1/2)/s)t_{i1}+\sin(2\pi(j+1/2)/s)t_{i2}\). The independent reference maximized the load multiplier over Cartesian contact forces \(f_i\in\mathbb R^3\):
\begin{equation}
\begin{aligned}
 q_k^{\mathrm{cart}}={}&\max_{\lambda\ge0,\,\{f_i\}}\lambda,\\
 &\sum_i\begin{bmatrix}f_i\\(p_i-c)\times f_i/\rho\end{bmatrix}
       =\lambda\ell_k,\qquad
   \sum_i n_i^\top f_i\le1,\\
 &n_i^\top f_i\ge0,\qquad
   v_{ij}^\top f_i\le\mu\cos(\pi/s)n_i^\top f_i
   \quad\text{for every }i,j.
\end{aligned}
\label{eq:cartesian-reference}
\end{equation}
The face inequalities describe the same inscribed friction polygon as the cone edges in \eqref{eq:contact-wrench}. The reference calculation constructed its equilibrium matrices and polygon constraints independently of the production wrench builder. A separate affine-rank and strict-interior calculation applied the zero-without-force-closure convention before comparing \(\min_k q_k^{\mathrm{cart}}\) with \(Q\). This distinction is necessary because a contact set can support an individual load direction while lacking six-dimensional force closure.

We used contact sets from the six illustrated hand models, two analytic contact arrangements, and twelve seeded eight-contact arrangements. Four friction coefficients, twelve unit balancing-force directions, and two application points gave 1,920 directional capacity comparisons. The friction polygon had eight edges, and the total normal-force budget was one. Eighteen center-applied axis loads on the symmetric analytic arrangement had known unit capacity. The coordinate comparisons varied the common torque length from 1 to 1,000~mm, converted all lengths consistently between units, rotated the physical contact polygons with their frames, and shifted the moment origin in both the contact and task wrenches. Rotating the existing polygon preserved its physical edge directions during the frame comparison. Boundary, outside-hull, and rank-deficient wrench sets supplied additional strict-closure cases.

The coefficient program agreed with the independent Cartesian formulation in all 1,920 directional comparisons, with a maximum absolute capacity discrepancy of \(2.73\times10^{-10}\) (Table~\ref{tab:numerical-validation}). All eighteen analytic unit-capacity comparisons met the same tolerance. Across the 960 coordinate comparisons, the largest change was \(1.30\times10^{-13}\). Capacity agreement was assessed with absolute tolerance \(2\times10^{-7}\) plus relative tolerance \(2\times10^{-6}\). The largest equilibrium residual of the Cartesian reference was \(9.54\times10^{-14}\) under the unit normal-force budget. The rank-qualified min-weight and Ferrari--Canny predicates correctly distinguished all 88 interior, boundary, outside, and lower-dimensional cases.

\begin{table}[t]
\centering
\caption{\scshape Numerical Wrench Validation. Bold counts met their comparison criteria. Maximum discrepancies are absolute score differences, except the monotonicity row (largest decrease) and starred scale row. The latter measures the error of capacity times load-magnitude factor from one for positive analytic capacities, and absolute capacity for zero capacities. Closure cases were checked with both min-weight and Ferrari--Canny predicates.}
\label{tab:numerical-validation}
\small
\begin{tabular}{lrr}
\toprule
\rowcolor{TableHeader}
Quantity & Cases & Maximum discrepancy \\
\midrule
Task-wrench capacity & \mintcell{1,920} & \(2.73\times10^{-10}\) \\
Analytic unit capacity & \mintcell{18} & \(2.00\times10^{-10}\) \\
Coordinate transformations & \mintcell{960} & \(1.30\times10^{-13}\) \\
Friction monotonicity & \mintcell{1,440} & \(1.00\times10^{-10}\) \\
Strict closure classification & \mintcell{88} & --- \\
Scaled min-weight classification & \mintcell{448} & --- \\
Scaled directional capacity & \mintcell{9,408} & \(2.22\times10^{-16}\)\(^{*}\) \\
\bottomrule
\end{tabular}
\end{table}

Coordinate equilibration was examined over 448 constructed wrench polytopes with known interior, boundary, outside, or lower-dimensional origin geometry. Axis scales and task magnitudes ranged from \(10^{-12}\) to \(10^{12}\), including duplicated wrench columns. All 448 min-weight closure classifications and 9,408 directional-capacity comparisons met their analytic criteria. The largest scaled capacity discrepancy was \(2.22\times10^{-16}\), and the largest equilibrium residual in the equilibrated linear programs was \(4.44\times10^{-16}\). These comparisons evaluate the numerical coordinate and load scaling independently of the contact geometries used in the primary capacity study.

\subsection{Python and ROS Interface Latency}\label{sec:ros-benchmark}

We compared the direct Python binding and the GraspIt! ROS~1 service interface \citep{graspit_interface}, with default TCPROS connections and persistent service clients as distinct conditions. For the interface benchmark, we used GraspIt! engine revision \texttt{5f254577199fe0cab8c2ea094d32a66c118f2ba2}. All three interfaces were evaluated with a 50~mm cube and the bundled L6 model with eleven independent coordinates. The palm and cube rotations were identity, the cube center was (40.1, 10.3, 93.1)~mm in the palm frame \(\{H\}\), and every closure began from zero independent coordinates. Rendering, the finger-closing speed factor of one, and the option to continue closing after first contact were held fixed across interfaces.

We timed six process launches per condition on an Intel Core i7-13700F under Ubuntu 20.04 with ROS Noetic 1.17.4 and Python 3.12.10. Each launch used a four-CPU container with 6~GiB of memory and a software-rendered virtual display. The ROS master and display were running before startup timing, which spanned process launch through world loading and initial-state retrieval. After five warm-up cycles per launch, we ran 20 timed reset--close--score--state cycles and 100 object-pose reads, giving 120 cycles and 600 pose queries per condition. Native radius and volume were obtained through two binding calls or one ROS quality request. We ran the conditions sequentially and rotated their order across launches. These timings measure the complete operation paths, including native computation, accessors, and event handling.

\begin{table}[t]
\centering

\caption{\scshape Python and ROS Interface Timings. Latencies are medians [Q1, Q3], with units given per operation. The complete-cycle 95th percentile is reported separately. Throughput is completed cycles divided by total measured cycle time. Bold entries identify the lowest value of each latency statistic and the highest throughput. Italic entries identify the highest latency and lowest throughput.}
\label{tab:ros-benchmark}
\small
\begin{tabular}{lrrr}
\toprule
\rowcolor{TableHeader}
Operation & Direct Python & ROS~1 & Persistent ROS~1 \\
\midrule
Startup (s) & \mintcell{1.1226 [1.1198, 1.1256]} & 1.2565 [1.2552, 1.2593] & \salmoncell{1.2579 [1.2569, 1.2583]} \\
Object pose (\(\mu\)s) & \mintcell{2.168 [2.140, 2.206]} & \salmoncell{661.7 [612.2, 712.0]} & 12.68 [12.28, 14.35] \\
Autograsp (ms) & \mintcell{74.65 [74.47, 74.89]} & 116.54 [114.73, 118.06] & \salmoncell{117.24 [116.08, 118.46]} \\
Wrench scores (ms) & \salmoncell{75.01 [74.88, 75.16]} & \mintcell{43.53 [43.31, 43.90]} & 43.77 [43.47, 44.04] \\
Complete cycle (ms) & \mintcell{149.85 [149.51, 150.24]} & \salmoncell{167.87 [165.09, 169.92]} & 167.79 [166.26, 170.27] \\
Cycle, 95th percentile (ms) & \mintcell{156.70} & 177.43 & \salmoncell{177.83} \\
Throughput (cycles/s) & \mintcell{6.64} & 5.95 & \salmoncell{5.92} \\
\bottomrule
\end{tabular}
\end{table}

All 360 measured closures returned 13 contacts and identical closed coordinates. Native Ferrari--Canny quality was \(\varepsilon_{\mathrm{GI}}=0.0317\), and wrench-space volume was 0.0902; the absolute output differences were below \(10^{-9}\). The native and ROS contact queries returned points on opposing contact surfaces. After transformation into the cube frame and one-to-one minimum-distance assignment, the largest paired gap was 0.188~mm.

\begin{figure}[!htbp]
\centering
\includegraphics[width=\linewidth]{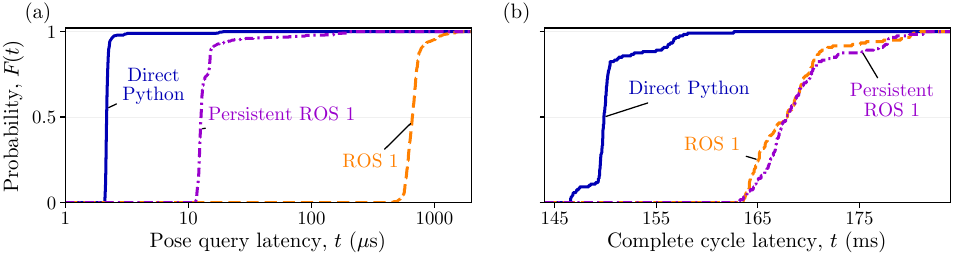}
\caption{\textbf{Latency by Interface and Workload.} Empirical cumulative distributions for direct Python, default-connection ROS~1, and persistent ROS~1. (a) Object-pose query latency over 600 warm reads per condition, on a logarithmic time axis. (b) Complete-cycle latency over 120 cycles per condition. Each cycle resets the hand, closes it, computes wrench scores, and reads the resulting state.}
\label{fig:interface-latency}
\end{figure}

Figure~\ref{fig:interface-latency} distinguishes the pose-read and complete-cycle workloads. The direct pose-read median was 2.17~\(\mu\)s, compared with 12.7~\(\mu\)s for persistent ROS. For a full cycle, the median decreased from 167.79 to 149.85~ms, a difference of 17.9~ms, or 10.7\% relative to persistent ROS. Closure and scoring accounted for most of the elapsed time. The operation-specific timings in Table~\ref{tab:ros-benchmark} also show why the complete-cycle difference cannot be interpreted as an isolated transport measurement: the quality calls differed in their accessor and refresh paths.

\subsection{Runtime and Memory Scaling}\label{sec:runtime-scaling}

The cost of repeated closure and numerical scoring determines the practical size of pose--friction studies. We measured a workload matrix with the RealHand L6, Barrett, and LEAP models and three object geometries: a 50~mm cube, a 50~mm sphere, and a cylinder of 44~mm diameter and 50~mm height. Each hand--object scene began without interpenetration. Direct Python, default ROS~1 connections, and persistent ROS~1 connections used the same native engine, model assets, closing command, and requested outputs. Each condition had five process launches, three warm-up closure cycles, and ten warm-up pose queries. We rotated the interface order across repetitions. Four reset--close--score--state cycles formed each primary workload, with further workloads varying the number of repeated closures, pose perturbations, or numerical samples. The complete evaluation contained 235 process launches, including 975 measured closure cycles and 4,875 object-pose queries.

We ran the workloads on an Intel Core i7-13700F with a two-CPU quota, a 3~GiB memory limit, and one BLAS thread. The native and ROS clients used Python~3.8.10 in a ROS Noetic environment, and the numerical sampling used the Python~3.12 runtime with the evaluated numerical implementation. Both interface paths used the same engine and Qhull compatibility correction. Closure rendering remained enabled in the engine. The direct score getters refreshed their values individually, while the ROS service obtained the quality pair together. The reported times therefore describe the complete requested operation paths, including their scoring and event handling.

Table~\ref{tab:multi-hand-runtime} reports median cycle costs and interquartile ranges across the five launches. Joint coordinates and object poses agreed in all nine primary scenes. Six scenes also satisfied the fixed output criteria: absolute quality differences below \(10^{-8}\), equal contact counts, and distances below 1~mm between opposing hand-side and object-side contact points after one-to-one pairing across interfaces. Across these six scenes, the ratio of ROS median elapsed time to direct-Python median elapsed time ranged from 1.57 to 4.49 over the two ROS connection modes. The L6 cube scene had contact-count variation between process instances, and the Barrett cube and cylinder scenes had volume variation beyond the fixed tolerance. The same volume variation occurred across process launches within an interface. In untimed controls, we kept a primitive wrench set fixed and changed its column order. The returned native volume varied across these permutations, although the geometric wrench hull was unchanged. We retained every launch and reported absolute costs for all scenes, with elapsed-time ratios confined to scenes satisfying the output criteria.

\begin{table}[t]
\centering
\caption{\scshape Multi-Hand Interface Costs. Values are median cycle latency in milliseconds, with interquartile range in parentheses, over five process launches per condition. Each launch measured four reset--close--score--state cycles. A dagger identifies contact or volume variation beyond the fixed output tolerances. Elapsed-time ratios were computed only for checked rows. Bold entries identify lower costs within these output-agreement rows.}
\label{tab:multi-hand-runtime}
\small
\setlength{\tabcolsep}{4pt}
\begin{tabular}{lrrrc}
\toprule
\rowcolor{TableHeader}
Scene & Direct Python & ROS~1 & Persistent ROS~1 & Agreement \\
\midrule
RealHand L6 / Cube & 149.1 (0.79) & 180.3 (7.60) & 176.5 (1.17) & \(\dagger\) \\
RealHand L6 / Sphere & \mintcell{73.4 (0.58)} & 119.4 (1.73) & 115.4 (0.86) & \(\checkmark\) \\
RealHand L6 / Cylinder & \mintcell{43.7 (0.63)} & 100.8 (1.84) & 94.8 (1.89) & \(\checkmark\) \\
Barrett / Cube & 15.6 (0.69) & 38.0 (1.23) & 36.2 (1.52) & \(\dagger\) \\
Barrett / Sphere & \mintcell{8.24 (0.08)} & 34.1 (0.15) & 32.6 (1.86) & \(\checkmark\) \\
Barrett / Cylinder & 24.6 (1.75) & 43.5 (1.06) & 42.7 (2.78) & \(\dagger\) \\
LEAP / Cube & \mintcell{53.7 (0.73)} & 226.7 (1.31) & 221.5 (0.67) & \(\checkmark\) \\
LEAP / Sphere & \mintcell{39.0 (0.38)} & 175.1 (2.60) & 171.2 (0.01) & \(\checkmark\) \\
LEAP / Cylinder & \mintcell{65.2 (0.38)} & 198.3 (2.75) & 193.7 (0.88) & \(\checkmark\) \\
\bottomrule
\end{tabular}
\end{table}

\begin{figure}[t]
\centering
\includegraphics[width=\linewidth]{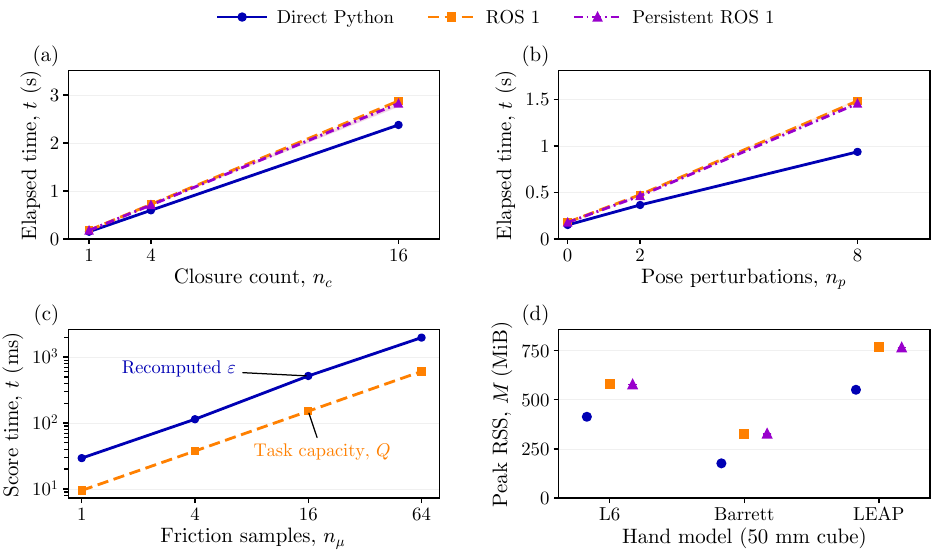}
\caption{\textbf{Closure, Numerical, and Memory Costs.} Medians and interquartile ranges over five process launches. (a) Repeated L6 cube closures. (b) Nominal pose with two or eight perturbations, including collision handling. (c) Friction-node sweeps for a fixed thirteen-contact set, with separate timings for matrix construction, recomputed Ferrari--Canny quality \(\varepsilon\), and task-wrench capacity \(Q\) on logarithmic axes. (d) Peak sampled process-tree resident memory for cube scenes. The legend in (a) applies to (b) and (d). Lines connect measured workload sizes.}
\label{fig:runtime-scaling}
\end{figure}

Figure~\ref{fig:runtime-scaling} separates repeated native closure, pose-dependent work, numerical scoring, and resident memory. For the L6 cube, increasing the closure count from one to sixteen increased the direct workload median from 0.150 to 2.38~s. Pose perturbations used a common seed with 1~mm translation standard deviation and 0.25~degrees rotation standard deviation. Every assigned pose contributed to the measured cost. Four of the nine poses in the largest workload entered the collision branch in every interface, so the pose curves include collision handling and state queries for every assigned pose. For numerical sampling, we held a thirteen-contact set fixed and varied the friction-node count. Both numerical calculations used eight cone edges per contact and a 100~mm torque length. At 64 nodes, matrix construction took 105~ms, recomputed Ferrari--Canny evaluation took 1.98~s, and task-wrench capacity over six unit Cartesian force directions through the center of mass took 607~ms. These costs distinguish repeating contact formation under pose changes from evaluating friction changes on fixed contacts.

We measured resident memory as the maximum sampled sum of process RSS over each complete process tree, including the client, engine, display, ROS master, and rosout. The sampler targeted a 10~ms interval, with a maximum observed interval of 41.1~ms. Shared resident pages contribute to each process's RSS, so the reported sum describes the process-tree footprint. The largest sampled total was 768~MiB. Median resident memory stayed within 414--416~MiB across one to sixteen repeated L6 closures, and within 82.4--95.7~MiB across one to 64 numerical friction samples. For the cube scenes in Figure~\ref{fig:runtime-scaling}(d), direct-Python median peaks ranged from 177 to 551~MiB, while the ROS deployments ranged from 328 to 766~MiB.

\subsection{Grasp Planning Across Multiple Embodiments}\label{sec:embodiment-replay}
\begin{figure}[!htbp]
\centering
\input{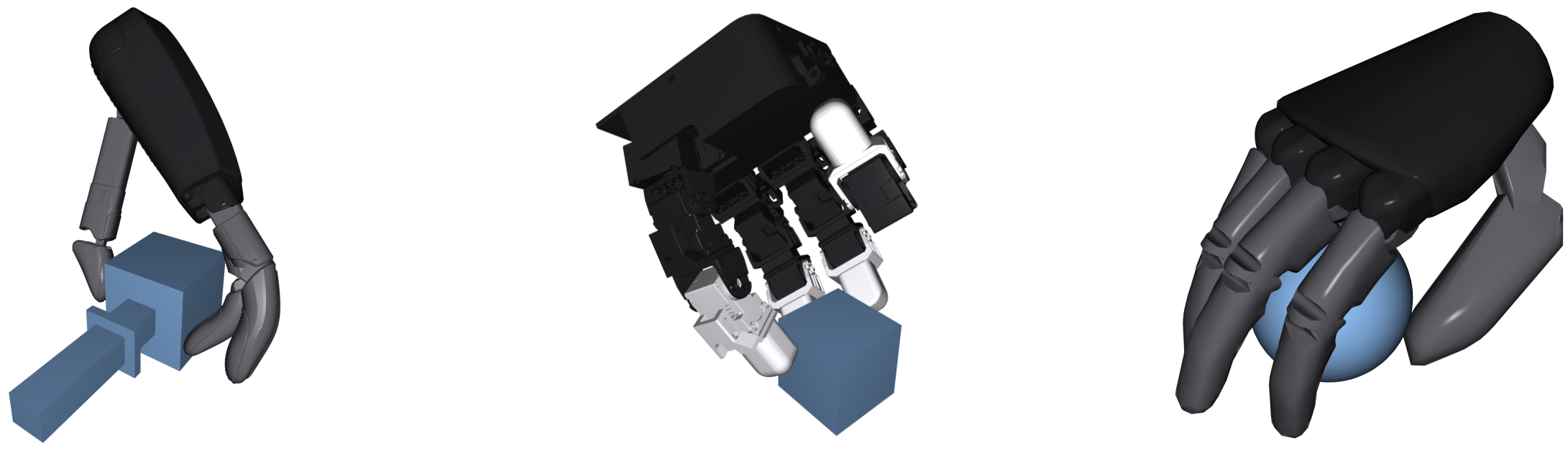}
\caption{\textbf{Grasp Commands Across Three Hand Embodiments.} (a) Key grasp reconstructed for bimanual insertion. (b) Final cube hold with 53.6~mm minimum rise. (c) Final sphere hold with 57.8~mm minimum rise. The hold configurations are expressed in their payload frames. Subcaptions report native GraspIt! Ferrari--Canny quality \(\varepsilon_{\mathrm{GI}}\), evaluated with each hand's corresponding object model at the planned closure.}
\label{fig:three-embodiments}
\end{figure}

Using a planned grasp in task simulation requires reconstruction of its palm pose and joint coordinates in the arm--hand model. For the RealHand L6, LEAP, and Ability examples, we reconstructed commands using the DexReach arm--hand models \citep{enweremGraspDistanceFields2026}. The object-relative palm pose and independent coordinates specify the grasp; the hand manifest expands the independent coordinates into source hinge angles. An external arm planner supplies the motion needed to reach and carry the payload. Figure~\ref{fig:three-embodiments} shows the selected L6 key grasp and final payload-relative configurations from the LEAP and Ability holds.

For the illustrated lifts, we used a 50~mm cube with ZArm--LEAP and a 60~mm sphere with Franka--Ability. Both payloads had mass 100~g and sliding friction 0.7. The arm moved from an open standoff through closure and a 60~mm upward trajectory. At the sampled observations during each illustrated two-second hold, at least three distal fingers were in contact. For Ability, three load-bearing distal fingers were observed at all 2,000 hold samples in each of three simulator resets. These contact histories and minimum rises describe the illustrated executions. We specified the arm trajectories and dynamic controllers in the task simulator.

The examples relate numerical grasp analysis to command reconstruction and lift execution. Native contacts and the joint map identify the closed hand configuration, and the numerical wrench calculation exposes the scoring convention. The five-record workflow in Section~\ref{sec:runnable-workflow} demonstrates reconstruction from an installed package and explicit assets. Its kinematic residuals measure restoration of the planned command, while the lift observations measure payload retention during motion. The contact and coordinate representations allow both quantities to be traced to the corresponding model and command.

\section{Conclusion}\label{conclusion}

We introduced \texttt{graspit-py}, a Python interface to GraspIt! that exposes closure and contact quantities for evaluating grasp commands under pose and friction uncertainty. Grasp records preserve the accepted object-relative placement, closed independent coordinates, nominal contacts, and sampled scores, connecting numerical analysis with command reconstruction. The exporter derives the affine joint correspondence, including reference-angle offsets, and preserves individual collision components. The link comparisons established micrometer-scale correspondence across all five evaluated hand models. The controlled ablations quantify the displacement caused by omitting a nonzero reference contribution and the occupied regions introduced by combining collision components.

We evaluated numerical consistency and computational cost across the interface's grasp-analysis operations. Task-wrench capacity is evaluated through coefficient-allocation linear programs with a common normal-force budget and equilibrated wrench coordinates. Agreement with an independent Cartesian-force formulation, coordinate transformations, and analytic scale cases establishes numerical consistency over the evaluated inputs. The runtime study quantifies repeated closure, pose perturbation, friction sampling, and process-tree resident memory across three hands and three object geometries. Output-agreement checks identify the scenes supporting elapsed-time comparisons. The executable workflow and six-hand examples demonstrate the interface from model conversion through numerical scoring and reconstruction. Future work includes a browser-based viewer for inspecting saved grasp geometry, coordinate frames, contacts, and quality samples.

\section{Data Availability}\label{sec:artifact}
The source code is available in \href{https://github.com/coenwerem/graspit-py}{\texttt{graspit-py}}, and the \href{https://clintonenwerem.com/graspit-py/}{manual} describes installation, supported hand models, and numerical interfaces. The public \href{https://github.com/coenwerem/graspit-py/tree/main/examples/l6_workflow}{L6 example} provides executable export, planning, and reconstruction programs with their search configuration. Hand-mesh sources and licenses are listed in the \href{https://clintonenwerem.com/graspit-py/hands.html}{hand-model guide}.

The \href{https://github.com/coenwerem/graspit-py/releases/download/v0.2.0/interface-evaluation-data.zip}{interface-evaluation archive} contains supporting geometry measurements, interface-timing samples, grasp configurations, and analysis programs \citep{enwerem2026interfaceData}.

The journal supporting material contains the exporter-ablation and scaling measurements, the Cartesian-force reference, the numerical implementation used for the extended evaluation, and the corresponding analysis programs.

\section*{Funding}
This work was supported by Grant No. GR100333 and a grant from Northrop Grumman Corporation.

\section*{Conflict of Interest}
The authors declare no competing interests.

\FloatBarrier
\setlength{\bibsep}{1pt}
\interlinepenalty=10000

\end{document}